\documentclass[10pt,twocolumn,letterpaper]{article}

\usepackage{iccv}
\usepackage{times}
\usepackage{epsfig}
\usepackage{graphicx}
\usepackage{amsmath}
\usepackage{amssymb}

\usepackage{accsupp}[axessibility]  
\usepackage{multirow}

\usepackage[breaklinks=true,bookmarks=false]{hyperref}

\iccvfinalcopy 

\def\iccvPaperID{****} 

\ificcvfinal\fi

\begin{document}

\title{Digging into Uncertainty in Self-supervised Multi-view Stereo}

\author{
Hongbin Xu\textsuperscript{\rm 1, \rm 2},
Zhipeng Zhou\textsuperscript{\rm 4},
Yali Wang\textsuperscript{\rm 1},
Wenxiong Kang\textsuperscript{\rm 2, \rm 5},
Baigui Sun\textsuperscript{\rm 4},
Hao Li\textsuperscript{\rm 4},
Yu Qiao\textsuperscript{\rm 1,3}\thanks{Corresponding author.}\\

\textsuperscript{\rm 1}ShenZhen Key Lab of Computer Vision and Pattern Recognition,\\
Shenzhen Institute of Advanced Technology, Chinese Academy of Sciences\\
\textsuperscript{\rm 2}South China University of Technology,
\textsuperscript{\rm 3}Shanghai AI Laboratory,
\textsuperscript{\rm 4}Alibaba Group,
\textsuperscript{\rm 5}Pazhou Laboratory\\
{\tt\small hongbinxu1013@gmail.com}\quad
{\tt\small yu.qiao@siat.ac.cn}
}

\maketitle
\ificcvfinal\thispagestyle{empty}\fi

\begin{abstract}
Self-supervised Multi-view stereo (MVS) with a pretext task of image reconstruction has achieved significant progress recently.
However, previous methods are built upon intuitions, lacking comprehensive explanations about the effectiveness of the pretext task in self-supervised MVS.
To this end, we propose to estimate epistemic uncertainty in self-supervised MVS, accounting for what the model ignores.
Specially, the limitations can be categorized into two types: ambiguious supervision in foreground and invalid supervision in background.
To address these issues, we propose a novel Uncertainty reduction Multi-view Stereo (U-MVS) framework for self-supervised learning.
To alleviate ambiguous supervision in foreground, we involve extra correspondence prior with a flow-depth consistency loss.
The dense 2D correspondence of optical flows is used to regularize the 3D stereo correspondence in MVS.
To handle the invalid supervision in background, we use Monte-Carlo Dropout to acquire the uncertainty map and further filter the unreliable supervision signals on invalid regions.
Extensive experiments on DTU and Tank\&Temples benchmark show that our U-MVS framework\footnote{Code: \url{https://github.com/ToughStoneX/U-MVS}} achieves the best performance among unsupervised MVS methods, with competitive performance with its supervised opponents.
\end{abstract}

\section{Introduction}
\label{sec:introduction}

Multi-view stereo (MVS) \cite{seitz2006comparison} is a fundamental computer vision problem which aims to recover 3D information from multiple images on different views. 
Standing on the shoulder of giants in traditional methods \cite{galliani2015massively,schonberger2016structure}, recent learning-based methods \cite{yao2018mvsnet,yao2019recurrent} have extended the MVS pipeline to deep neural networks, achieving state-of-the-art performance in several benchmarks \cite{aanaes2016large,Knapitsch2017}.
However, the fully supervised learning paradigm suffers from the non-negligible problem of requiring tedious and expensive procedures for collecting ground truth depth annotations.
Hence, it leads the community to consider competitive alternative of learning-based approaches which require fewer labels.

\begin{figure}
\begin{center}
  \includegraphics[width=\linewidth]{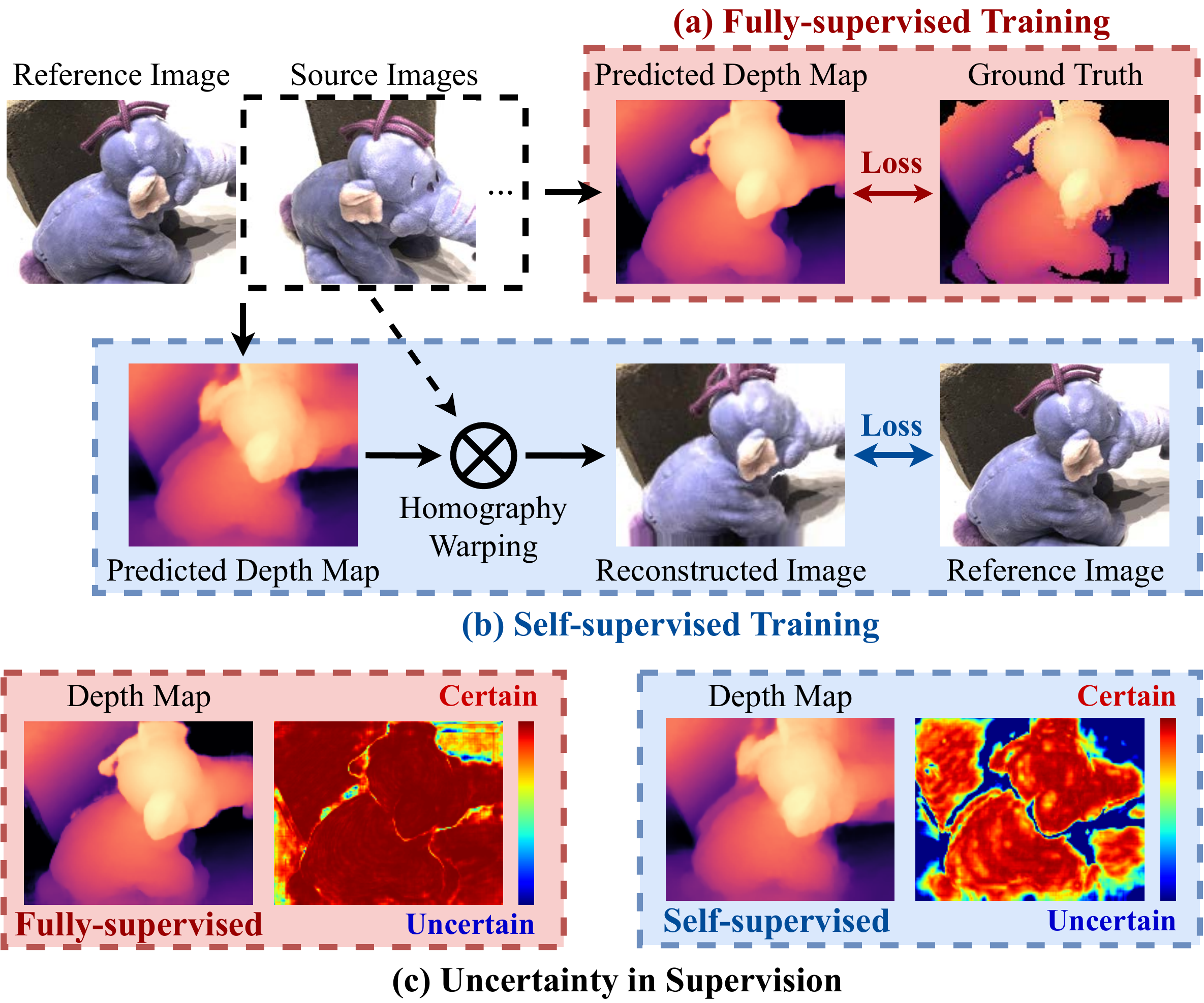}
\end{center}
    \vspace{-0.3cm}
    \caption{Illustration of the effectiveness of fully-supervised and self-supervised training in learning-based MVS via the visualization of \emph{uncertainty in supervision}.}
    \vspace{-0.3cm}
\label{fig1}
\end{figure}

\begin{figure*}[t]
\begin{center}
\includegraphics[width=0.9\linewidth]{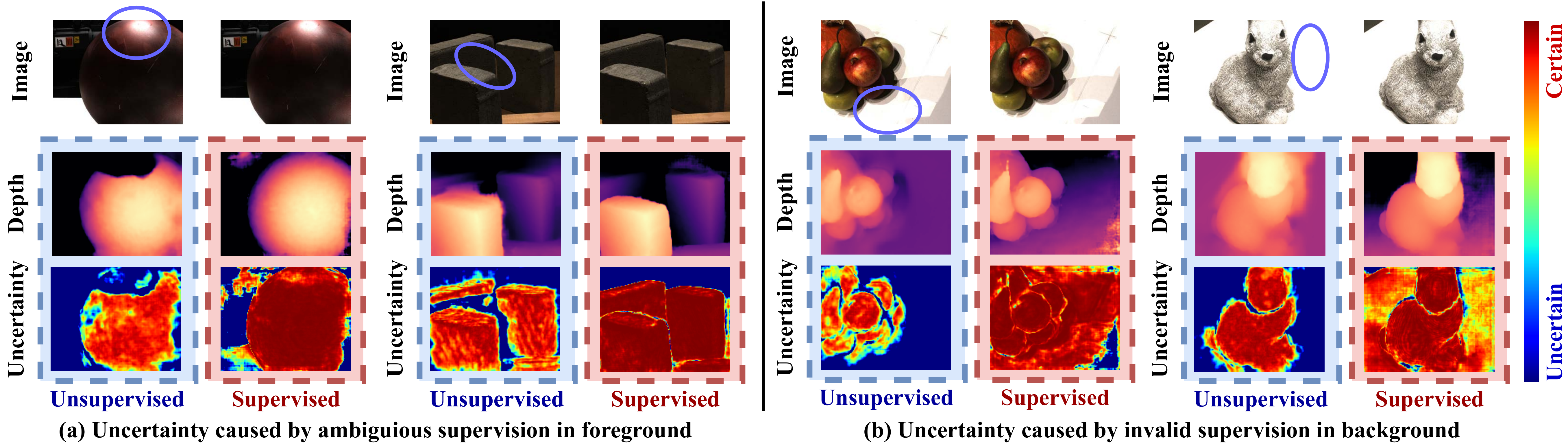}
\end{center}
    \vspace{-0.3cm}
    \caption{Visualization of epistemic uncertainty in fully-supervised and self-supervised MVS.}
    \vspace{-0.3cm}
\label{fig2}
\end{figure*}

A prominent and appealing trend is to construct a self-supervised MVS task \cite{godard2017unsupervised, khot2019learning, dai2019mvs2, huang2020m}, which further transforms the original depth estimation problem as RGB image reconstruction problem.
However, previous methods are merely built upon intuitive motivations, lacking comprehensive explanations about which image regions such self-supervision signal can effectively work for multi-view depth estimation.
For fully-supervised MVS (Fig. \ref{fig1}(a)), the regions where supervision exists are explicit, if given the ground truth depth map.
Whereas, for self-supervised MVS shown in Fig. \ref{fig1}(b), the pretext task of image reconstruction actually provides indistinct supervision based on color similarity, which is agnostic to the exact presence of supervision in depth estimation.
Hence, to provide a direct proof of the effectiveness in supervision, we utilize Monte-Carlo Dropout \cite{kendall2017uncertainties} to visualize the epistemic uncertainty for a comprehensive insight (Fig. \ref{fig1}(c)).
In Bayesian modeling \cite{der2009aleatory}, the epistemic uncertainty inherently reflects what the supervision ignores.

\emph{What can we know from uncertainty?}
In Fig. \ref{fig2}, we provide a direct comparison of uncertainty between fully-supervised and self-supervised MVS to explicitly understand \emph{what factors may lead to the failure of self-supervision}.
It is found that the uncertain regions in self-supervision are more than the ones in fully-supervised training.
Revisiting the premise of self-supervision as an image reconstruction task, the problem can be distinguished into two groups:
(1) \emph{Ambiguous supervision in foreground} (Fig. \ref{fig2}(a)).
Under the influence of unexpected factors such as color variation and occlusion in the foreground object, the pretext task of image reconstruction is inconsistent with the photometric consistency and unable to reflect the correct depth information.
(2) \emph{Invalid supervision in background} (Fig. \ref{fig2}(b)).
The textureless background provides no effective clues for depth estimation task, which is ignored in fully-supervised training.
Whereas, the pretext task of image recostruction takes the whole image including the textureless backgrounds into consideration, involving invalid supervisions and oversmoothing the self-supervised results.

\emph{How to handle these uncertainties?}
To address these problems, we propose a novel Uncertainty reduction Multi-view Stereo framework U-MVS for self-supervised learning.
It mainly consists of two distinct designs as follows:
(1) To handle \emph{ambiguous supervision in foreground}, we aim to append extra prior of correspondence to strengthen the reliability of self-supervision, and introduce a new multi-view flow-depth consistency loss.
The intuition is that the dense 2D correspondence of optical flow can be utilized to regularize the 3D stereo correspondence in self-supervised MVS.
A differentiable Depth2Flow module is proposed to convert the depth map to virtual optical flow among views and the RGB2Flow module unsupervisedly predict the optical flow from corresponding views.
Then the virtual flow and the real flow are enforced to be consistent.
(2) To handle \emph{invalid supervision in background}, we suggest to filter the unreliable supervision signals on invalid regions, and propose an uncertainty-aware self-training consistency loss.
In a totally unsupervised setting, we firstly annotate the dataset with a self-supervisedly pretrained model, while acquiring the uncertainty map with Monte-Carlo Dropout.
Then the pseudo label filtered by the uncertainty map is used to supervise the model.
Random data-augmentations on the input multi-view images are appended to enforce the robustness towards disturbance on the areas with valid supervision.

In summary, our contributions are:
(1) We propose a novel self-supervised framework to handle the problems investigated from the visualization analysis about the uncertainty gap between supervised and self-supervised supervision signals.
(2) We propose a novel self-supervision signal based on the cross-view consistency of optical flows and depth maps among arbitrary views.
(3) We propose an uncertainty-aware self-training consistency loss for self-supervised MVS.
(4) Experimental results on DTU and Tanks\&Temples show that our proposed method can achieve competitive performance with its supervised counterparts with same backbones.

\begin{figure*}[t]
\begin{center}
\includegraphics[width=0.9\linewidth]{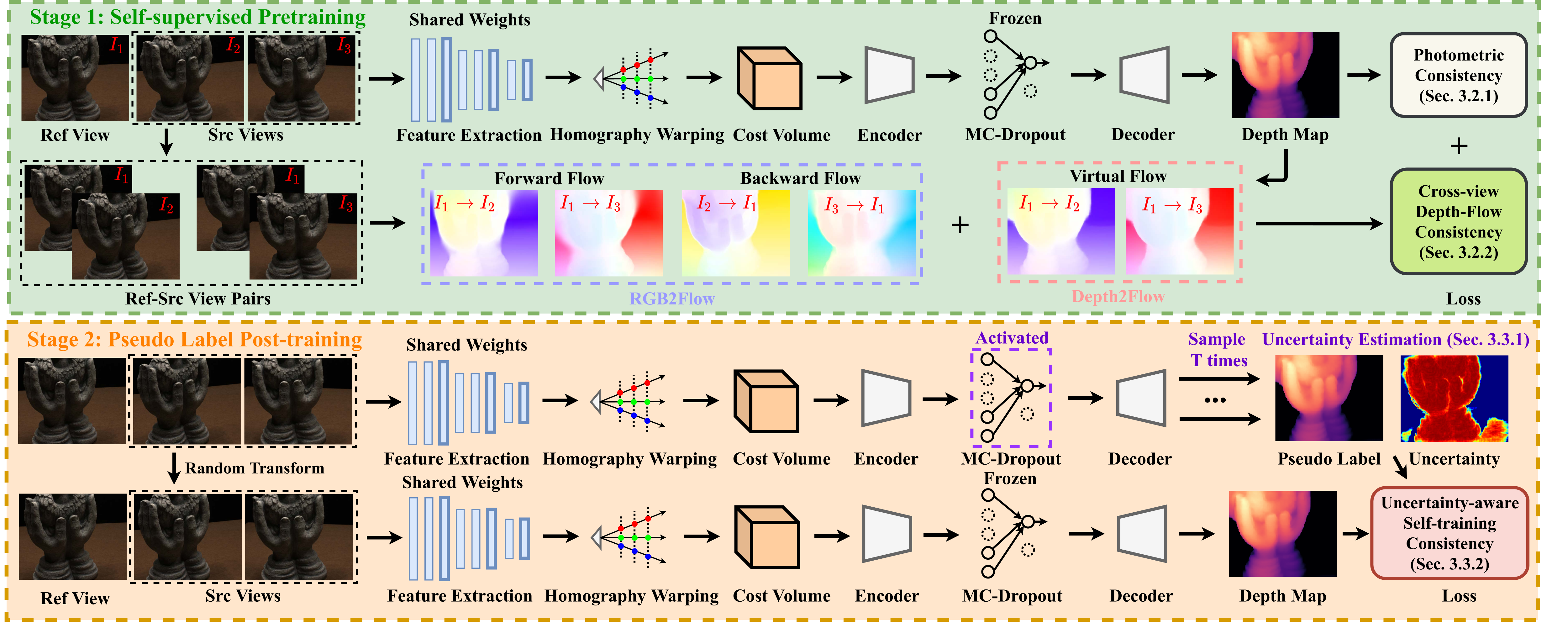}
\end{center}
    \vspace{-0.3cm}
    \caption{Illustration of our proposed self-supervised MVS framework U-MVS. ``MC-Dropout'' means Monte-Carlo Dropout.}
    \vspace{-0.3cm}
\label{fig3}
\end{figure*}

\section{Related Work}
\label{sec:related_work}

\noindent\textbf{Supervised Multi-view Stereo:}
With the flourishing of deep learning, convolutional neural networks (CNN) have now superseded classical techniques in Multi-view stereo.
MVSNet \cite{yao2018mvsnet} is a profound attempt that builds a standard MVS pipeline with end-to-end neural networks.
They utilize 3D CNN to regularize the cost volume from features of CNN and get the depth map based on the soft-argmin operation from the output volume.
Many efforts are further made to relieve the huge memory cost of cost volume.
R-MVSNet \cite{yao2019recurrent} replace the 3D convolution with recurrent convolutional GRU unit.
Many concurrent works are built on coarse-to-fine manner by separating the single cost volume regression into multiple stages, such as Fast-MVSNet \cite{yu2020fast}, UCS-Net \cite{cheng2020deep}, CVP-MVSNet \cite{yang2020cost} and CascadeMVSNet \cite{gu2020cascade}, achieving resounding success.

\noindent\textbf{Unsupervised / Self-supervised Multi-view Stereo:}
The burgeoning field of self-supervision \cite{godard2017unsupervised} provide a competitive alternative for amazing performance and requiring no ground truth data.
In Unsup\_MVS \cite{khot2019learning}, the predicted depth map and the input images are utilized to reconstruct the image on another view by homography warping, thus the photometric consistency is enforced to minimize the difference between the original and reconstructed images.
MVS$^2$ \cite{dai2019mvs2} predicts the per-view depth maps simultaneously and automatically infer the occlusion relationship among views.
M$^3$VSNet \cite{huang2020m} enforce the consistency between surface normal and depth map to regularize the MVS pipeline.
JDACS \cite{xu2021self} revisit the color constancy hypothesis of self-supervision and propose a unified framework to enhance the robustness of self-supervision signal towards the natural color disturbance in multi-view images.

\noindent\textbf{Uncertainty:}
The uncertainty \cite{der2009aleatory} in deep learning models for vision tasks can be classified as 
aleatoric uncertainty and epistemic uncertainty.
Aleatoric uncertainty captures the noise inherent in the training data, while epistemic uncertainty provides interpretation for the uncertainty in the model which can be remedied with enough data.
\cite{kendall2017uncertainties} study the benefits of modeling epistemic and aleatoric uncertainty in Bayesian deep learning models for vision tasks.
In this work, we aim to reject the unreliable pixels estimated by the epistemic uncertainty.
Similar idea also appears in \cite{mostegel2016using}.
Confidence estimation is applied in MVS to filter the unreliable predictions, such as \cite{kuhn2020deepcmvs,li2020confidence}.
UCS-Net \cite{cheng2020deep} progressively reconstruct high-resolution depth map with a coarse-to-fine manner. The depth hypothesis of each stage adapts to the uncertainties of previous per-pixel depth predictions.

\section{Method}
\label{sec:method}

In this section, we introduce the proposed self-supervised MVS framework, U-MVS.
As Fig. \ref{fig3} shows, the architecture of U-MVS is comprised of two stages: self-supervised pre-training stage and pseudo label post-training stage.
The backbone model (Sec. \ref{sec:method:backbone}) is firstly trained in the self-supervised pre-training stage (Sec. \ref{sec:method:self_supervised_pretraining}), and then trained in the pseudo label post-training stage (Sec. \ref{sec:method:pseudo_label_posttraining}).

\subsection{Backbone}
\label{sec:method:backbone}
Arbitrary MVS network can be utilized as backbone in our self-supervised MVS framework.
In default, the representative MVSNet \cite{yao2018mvsnet} is used.
The network extracts feature from $N$ input multi-view images and reprojects the feature maps in source views to the reference view by differentiable homography warping.
The variance of the feature maps are used to construct a cost volume and a 3D U-Net is utilized to regularize the volume.
Different from the standard 3D U-Net, we apply Monte-Carlo Dropout \cite{kendall2017uncertainties} on the bottleneck layer between the encoder and decoder, as shown in Fig. \ref{fig3}.
In default, the Monte-Carlo Dropout is frozen when predicting depth map.
It is only activated when estimating the uncertainty maps and pseudo labels.

\subsection{Self-supervised Pre-training}
\label{sec:method:self_supervised_pretraining}

The self-supervised pre-training stage contains two components of self-supervision loss: photometric consistency loss and cross-view depth-flow consistency loss.
In the photometric consistency loss, the images on the source views are utilized to reconstruct the image on the reference view via homography warping relationship determined by the predicted depth map.
As a solution to the \emph{ambiguous supervision in foreground}, we add an extra branch of depth-flow consistency to endow extra correspondence prior to the self-supervision loss.

\subsubsection{Photometric Consistency}
The core insight of photometric consistency \cite{barnes2009patchmatch} aims at minimizing the difference between the real image and the synthesized image from other views.
It is denoted that the first view is the reference view and the $j(2 \leq j \leq V)$-th view is one of the $V-1$ source views.
For a pair of multi-view images $(I_1, I_j)$, it is attached with the intrinsic and extrinsic camera parameters $([K_1, T_1], [K_j, T_j])$.
The output of a MVSNet backbone is the depth map $D_1$ on the reference view.
We can compute the corresponding point position of pixel $\hat{p}_i$ in the source view $j$ according to its position $p_i^j$ in the reference view. 
\begin{equation}
    D_j (\hat{p}_i^j) \hat{p}_i^j = K_j T_j (K_1 T_1)^{-1} D_1(p_i) p_i
    \label{eq1}
\end{equation}
where $i(1 \leq i \leq HW)$ represents the index of pixels in the images.
Since $D_j (\hat{p}_i^j)$ is a scale term in homogeneous coordinates, the $\hat{p}_i^j$ can be further described by the following equation:
\begin{equation}
    \hat{p}_i^j = \text{Norm} [D_j (\hat{p}_i^j) \hat{p}_i^j]
    \label{eq2}
\end{equation}
where $\text{Norm} ([x, y, z]^T) = [x/z, y/z, 1]^T$.

Then the synthesized image $\hat{I}_1^j$ from the source view $j$ to the reference view can be calculated via differentiable bilinear sampling \cite{jaderberg2015spatial}.
In Eq. \ref{eq1}, we can obtain a binary mask $M_j$ indicating the valid corresponding pixels of $I_j$ to the synthesized image $\hat{I}_1^j$.
In a self-supervised MVS system, all source views are warped into the reference view to calculate the photometric consistency loss:
\begin{equation}
    \small
    L_{pc} = \sum_{j=2}^{V} \frac{\| (I_1 - \hat{I}_1^j) \odot M_j \|_2 + \| (\nabla I_1 - \nabla \hat{I}_1^j) \odot M_j \|_2}{\| M_j \|_1}
    \label{eq3}
\end{equation}
where $\nabla I$ represents the gradient on $x$ and $y$ direction of image $I$ and $\odot$ means point-wise product.

\subsubsection{Cross-view Flow-Depth Consistency}
As discussed in Sec. \ref{sec:introduction}, one problem of basic self-supervised MVS is the \emph{ambiguous supervision in foreground}.
To handle this issue, we propose a novel flow-depth consistency loss to regularize the self-supervision loss.
The flow-depth consistency loss is comprised of two modules: RGB2Flow and Depth2Flow, as shown in Fig. \ref{fig3}.
In the Depth2Flow module, the predicted depth map is transformed as a virtual optical flow between the reference view and arbitrary source view.
The whole Depth2Flow module is differentiable, which can be plugged in the training framework.
In the RGB2Flow module, we use an unsupervised method \cite{liu2020learning} to predict the optical flow from corresponding reference-source view pairs.
The forward and backward flows obtained from RGB2Flow module are required to be consistent with the virtual flow calculated from Depth2Flow module.

\begin{figure}[t]
\begin{center}
  \includegraphics[width=0.9\linewidth]{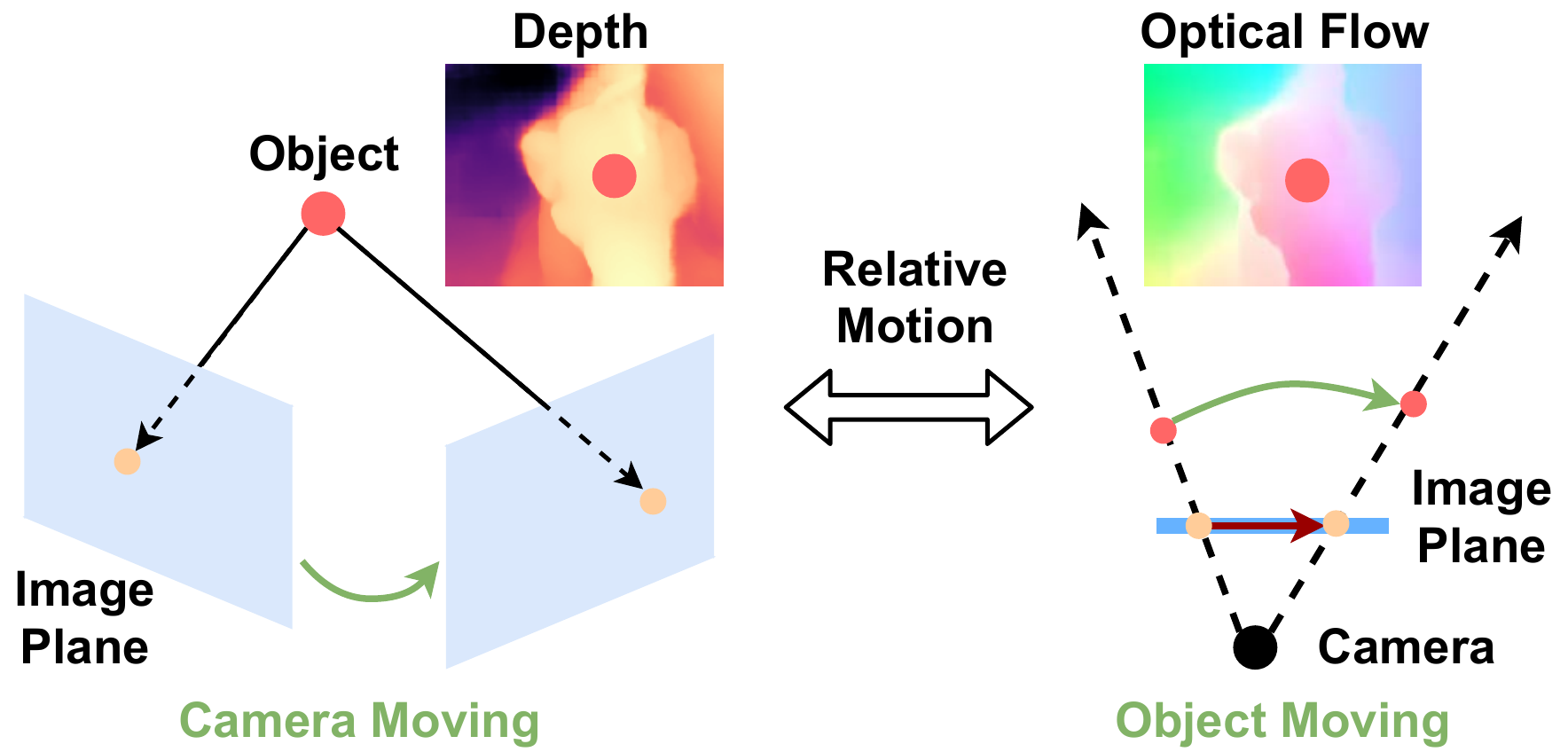}
\end{center}
    \vspace{-0.3cm}
    \caption{Intuition of Depth2Flow module. The relative motion of moving camera can be viewed as a special case of moving object represented by optical flow.}
    \vspace{-0.3cm}
\label{fig-virtual-flow}
\end{figure}

\textbf{Depth2Flow Module}:
In a standard MVS system, the cameras are moving around the target object with fixed position when collecting the multi-view images.
The relative motion of camera towards object can also be viewed as the motion of object towards a virtual camera with fixed position as shown in the Fig. \ref{fig-virtual-flow}.
In this way, the correspondence can be represented by a dense 2D optical flow between arbitrary views and should be consistent with the 3D correspondence determined by homography warping in real MVS system.
Given the definition of the aforementioned virtual optical flow: $\widehat{F}_{1j}(p_i) = \hat{p}_i^j - p_i$,
where $\widehat{F}_{1j}(p_{i})$ represent the optical flow between the corresponding point $p_i$ in the reference view and $\hat{p}_i^j$ in the sourve view $j$.
Considering the stereo correspondence defined in homography warping function (Eq. \ref{eq1} and Eq. \ref{eq2}):
\begin{equation}
    \widehat{F}_{1j} (p_i) = \text{Norm} [K_j T_j (K_1 T_1)^{-1} D_1(p_i) p_i] - p_i
    \label{eq4}
\end{equation}

With Eq. \ref{eq4}, the implicit correspondence modeled in the depth map can be explicitly transformed to the 2D correspondence of optical flow between the reference view and arbitrary source view $j$.
This operation is fully differentiable which can be inserted into the training framework, namely Depth2Flow module in Fig. \ref{fig3}.

\textbf{RGB2Flow module}:
We utilize a self-supervised method \cite{liu2020learning} to train a PWC-Net \cite{sun2018pwc} on the dataset from scratch.
All two-view pairs are enumerated among the provided multi-view pairs from the target MVS dataset.
After unsupervisedly pretrained on the MVS dataset, the PWC-Net is used to predict the optical flow in the RGB2Flow module.
As shown in Fig. \ref{fig3}, all two-view pairs combined with reference view and arbitrary source view are fed to RGB2Flow module.
The output includes the forward flow and backward flow among each of the two views.
The forward flow $F_{1j}$ models the projection from reference view to source view $j$.
In contrast, the backward flow $F_{j1}$ represents the optical flow from source view $j$ to reference view.

\textbf{Loss function}:
The predicted depth map $D_1$ can be converted to virtual cross-view optical flow $\widehat{F}_{1j}$ by Depth2Flow module.
The output of RGB2Flow module is forward flow $F_{1j}$ and backward flow $F_{j1}$, which should be consistent with the virtual flow $\widehat{F}_{1j}$.
For non-occluded pixels, the forward flow $F_{1j}$ should be the inverese of the backward flow $F_{j1}$.
To avoid learning incorrect deformation in occluded pixels, we mask out the occluded parts via the occlusion mask $O_{1j}$ infered by forward-backward consistency check \cite{meister2018unflow}:
\begin{equation}
    \small
    O_{1j} = \{ | F_{1j} + F_{j1} | > \epsilon \}
    \label{eq5}
\end{equation}
where the threshold $\epsilon$ is set to 0.5.
Then, the flow-depth consistency loss can be calculated:
\begin{equation}
    \small
    L_{fc} = \sum\limits_{i=1}^{HW}\min_{2\le j\le V} \frac{\| (F_{1j}(p_{i}) - \widehat{F}_{1j}(p_{i}))\cdot O_{1j}(p_{i}) \|_2}{ \sum_{i=1}^{HW} O_{1j}(p_{i}) }
    \label{eq6}
\end{equation}
At each pixel, instead of averaging the difference between $\widehat{F}_{1j}$ and $F_{1j}$ on all source views, we use the minimum error.
The minimum error was firstly introduced in \cite{godard2019digging} to reject occluded pixels in self-supervised monocular depth estimation.
Since the unsupervised RGB2Flow module may generate noisy predictions of optical flows, we utilize the minimum error to reject unreliable optical flow among views.

\subsubsection{Overall Loss}
In self-supervised pre-training stage, the overall loss is comprised of the photometric consistency loss $L_{pc}$ and the flow-depth consistency loss $L_{fc}$:
\begin{equation}
    \small
    L_{ssp} = L_{pc} + \lambda L_{fc}
    \label{eq7}
\end{equation}
where $\lambda$ is a weight to balance the scale of $L_{fc}$, which is set to 0.1 in default.
The flow-depth consistency loss aims to involve extra correspondence regularization to enhance the robustness of self-supervision loss towards real-world disturbances.

\subsection{Pseudo-Label Post-training}
\label{sec:method:pseudo_label_posttraining}

To handle the aforementioned problem of \emph{invalid supervision in background} in Sec. \ref{sec:introduction}, the invalid regions like textureless backgrounds are ignored in the pseudo-label post-training stage.
The uncertainty maps are estimated from the pretrained model in self-supervised pre-training stage via Monte-Carlo Dropout \cite{kendall2017uncertainties}. 
Then, the normalized uncertainty mask is adopted to filter the uncertain regions when calculating the uncertainty-aware self-training loss.

\subsubsection{Uncertainty Estimation}
In practice, the predictive uncertainty conveys skepticism about a model's output.
As discussed in Sec. \ref{sec:method:backbone}, Monte-Carlo Dropout \cite{kendall2017uncertainties} is applied to the bottleneck layers in the 3D U-Net of the backbone.
Following the modification strategy suggested by \cite{kendall2017uncertainties}, the original photometric consistency loss is modified as follows:
\begin{equation}
    \small
    L_{pc}^{\prime} = \sum_{j=2}^{V} \frac{\| (I_1 - \hat{I}_1^j) \odot M_j^{\prime} \|_2 + \| (\Delta I_1 - \Delta \hat{I}_1^j) \odot M_j^{\prime} \|_2}{\| M_j^{\prime} \|_2} + \frac{1}{2} \log \Sigma^2
    \label{eq8}
\end{equation}
where $\Sigma^2$ is the predicted variance of data noise, which is also called aleatoric uncertainty. Since $\Sigma^2$ is pixelwise uncertainty, the weighted mask is calculated by: $M_j^{\prime} = \frac{1}{2} \exp (-\log \Sigma^2) \odot M_j$. Then the self-supervision loss (Eq. \ref{eq7}) is further modified as follows:
\begin{equation}
    \small
    L_{ssp}^{\prime} = L_{pc}^{\prime} + \lambda L_{fc}
    \label{eq9}
\end{equation}

In our framework, a 6-layer CNN directly predicts the pixelwise aleatoric uncertainty $\Sigma^2$ from the input image.
Then, the model is pre-trained with modified loss $L_{ssp}^{\prime}$ in Eq. \ref{eq9} in the self-supervised pre-training stage.

Random Monte-Carlo Dropout \cite{kendall2017uncertainties} plays a role in sampling different model weights: $\mathbf{W}_{t} \sim q_{\theta}(\mathbf{W}, t)$, where $q_{\theta}(\mathbf{W}, t)$ is the random Dropout distribution in each sample.
Denote that in the $t$-th time of sampling, with a model weight of $\mathbf{W}_{t}$, the predicted depth map is $D_{1,t}$.
For our depth regression loss, the epistemic uncertainty is captured by the predictive variance of sampled depth maps:
\begin{equation}
    \small
    U_1 = \frac{1}{T} \sum_{t=1}^{T} D_{1,t}^2 - (\frac{1}{T} \sum_{t=1}^{T} D_{1,t})^2 + \frac{1}{T} \sum_{t=1}^{T} \sigma_t^2
    \label{eq10}
\end{equation}
where $(D_{1,t}, \sigma_t)_{t=1}^T$ is the sampled outputs with random Monte-Carlo Dropout. The mean prediction $\overline{D}_1 = \frac{1}{T} \sum_{t=1}^{T} D_{1,t}$ of the $T$ sampled outputs is treated as the pseudo label.

\subsubsection{Uncertainty-aware Self-training Consistency}

To alleviate the invalid supervision in background, we utilize the generated pseudo label and uncertainty map in the previous section to construct an uncertainty-aware self-training consistency loss.
A binary uncertainty mask $\widehat{U}_1$ can be calculated after normalizing the predicted uncertainty $U_1$:
\begin{equation}
    \small
    \widehat{U}_1 = \{exp(-U_1) > \xi\}
    \label{eq11}
\end{equation}
where $\xi=0.3$ is the threshold for calculating the binary mask $\widehat{U}_1$, which only retains the certain regions in self-supervison.
Then, the uncertainty-aware self-training consistency loss can be calculated:
\begin{equation}
    \small
    L_{uc} = \frac{\| (D_{1,\tau} - \overline{D}_1) \odot \widehat{U}_1 \|_2}{\| \widehat{U}_1 \|_1}
    \label{eq12}
\end{equation}
where $D_{1,\tau}$ represent the output of a randomly transformed multi-view images.
All images $(I_1, I_j)$ are transformed by data-augmentation operations $(\tau_1, \tau_j)$ randomly.
In the framework, we utilize standard data-augmentation operations \cite{xu2021self} which do not move pixels, such as color jitter, gamma correction, random crop and etc.
The output of the augmented input is required to be consistent with the pseudo label $\overline{D}_1$ on the valid regions filtered by $\widehat{U}_1$.

\begin{table}
\centering
\small
\begin{tabular}{l|l|p{22pt} p{22pt} p{22pt}}
    \hline 
    & Method & Acc. & Comp. & Overall\tabularnewline \hline \hline 
    \multirow{4}{*}{Geo.} & Furu \cite{furukawa2009accurate} & 0.613 & 0.941 & 0.777\tabularnewline
    & Tola \cite{tola2012efficient} & 0.342 & 1.190 & 0.766\tabularnewline
    & Camp \cite{campbell2008using} & 0.835 & 0.554 & 0.694\tabularnewline
    & Gipuma \cite{galliani2015massively} & 0.283 & 0.873 & 0.578\tabularnewline \hline 
    \multirow{9}{*}{Sup.} & Surfacenet \cite{ji2017surfacenet} & 0.450 & 1.040 & 0.745\tabularnewline
    & \emph{MVSNet} \cite{yao2018mvsnet} & 0.396 & 0.527 & 0.462\tabularnewline
    & CIDER \cite{xu2020learning} & 0.417 & 0.437 & 0.427\tabularnewline
    & P-MVSNet \cite{luo2019p} & 0.406 & 0.434 & 0.420\tabularnewline
    & R-MVSNet \cite{yao2019recurrent} & 0.383 & 0.452 & 0.417\tabularnewline
    & Point-MVSNet \cite{chen2019point} & 0.342 & 0.411 & 0.376\tabularnewline
    & Fast-MVSNet \cite{yu2020fast} & 0.336 & 0.403 & 0.370\tabularnewline
    & \emph{CascadeMVSNet} \cite{gu2020cascade} & 0.325 & 0.385 & 0.355\tabularnewline
    & UCS-Net \cite{cheng2020deep} & 0.330 & 0.372 & 0.351\tabularnewline
    & CVP-MVSNet \cite{yang2020cost} & 0.296 & 0.406 & 0.351\tabularnewline 
    & PatchMatchNet \cite{wang2020patchmatchnet} & 0.427 & 0.277 & 0.352\tabularnewline \hline 
    \multirow{8}{*}{UnSup.} & Unsup\_MVS \cite{khot2019learning} & 0.881 & 1.073 & 0.977\tabularnewline
    & MVS$^{2}$ \cite{dai2019mvs2} & 0.760 & 0.515 & 0.637\tabularnewline
    & M$^{3}$VSNet \cite{huang2020m} & 0.636 & 0.531 & 0.583\tabularnewline
    & Meta\_MVS \cite{mallick2020learning} & 0.594 & 0.779 & 0.687 \tabularnewline
    & JDACS\cite{xu2021self} & 0.571 & 0.515 & 0.543\tabularnewline
    & \emph{Ours+MVSNet} & 0.470 & 0.430 & 0.450\tabularnewline
    & \emph{Ours+CascadeMVSNet} & 0.354 & 0.3535 & 0.3537\tabularnewline
    \hline
\end{tabular}
\caption{Quantitative results on DTU evaluation benchmark. ``Geo.''/``Sup.''
/``Unsup.'' are respectively the abbreviation of Geometric/Supervised/Unsupervised
methods.}
\vspace{-0.2cm}
\label{tab1}
\end{table}

\subsection{Overall Training Procedure}

As shown in Fig. \ref{fig3}, our proposed self-supervised framework, U-MVS, is comprised of two stages: self-supervised pretraining and pseudo-label post-training.
In the first stage of self-supervised pre-training stage, the overall loss $L_{ssp}$ includes photometric consistency loss $L_{pc}$ and flow-depth consistency loss $L_{fc}$.
As suggested by \cite{kendall2017uncertainties}, the uncertainty is involved in $L_{ssp}$ to construct the modified loss $L_{ssp}^{\prime}$, which is used for training.
In the second stage of pseudo-label post-training stage, the pseudo-label and uncertainty map are estimated from the pre-trained model in previous stage via Monte-Carlo Dropout \cite{kendall2017uncertainties}. 
In the uncertainty-aware self-training loss $L_{uc}$, the pseudo-label filtered by the uncertainty map is used to supervised the model.
Standard random data-augmentation operations are involved in the post-training stage.

\section{Experiment}
\label{sec:experiment}


\noindent\textbf{Dataset:} DTU \cite{aanaes2016large} is a large-scale indoor MVS dataset collected by robotic arms.
For each of the 124 scenes in total, high-resolution images are captured on 49 different views with 7 controlled light conditions.
Tanks\&Temples \cite{Knapitsch2017} is a outdoor MVS dataset, which contains challenging realistic scenes.
Following the official split of MVSNet \cite{yao2018mvsnet}, we train the model on DTU training set and test on the DTU evaluation set.
To validate the generalization performance of the proposed method, we test it on the \emph{intermediate} and \emph{advanced} partition of Tanks\&Temples without any finetuning.

\begin{figure}[t]
\begin{center}
\includegraphics[width=0.9\linewidth]{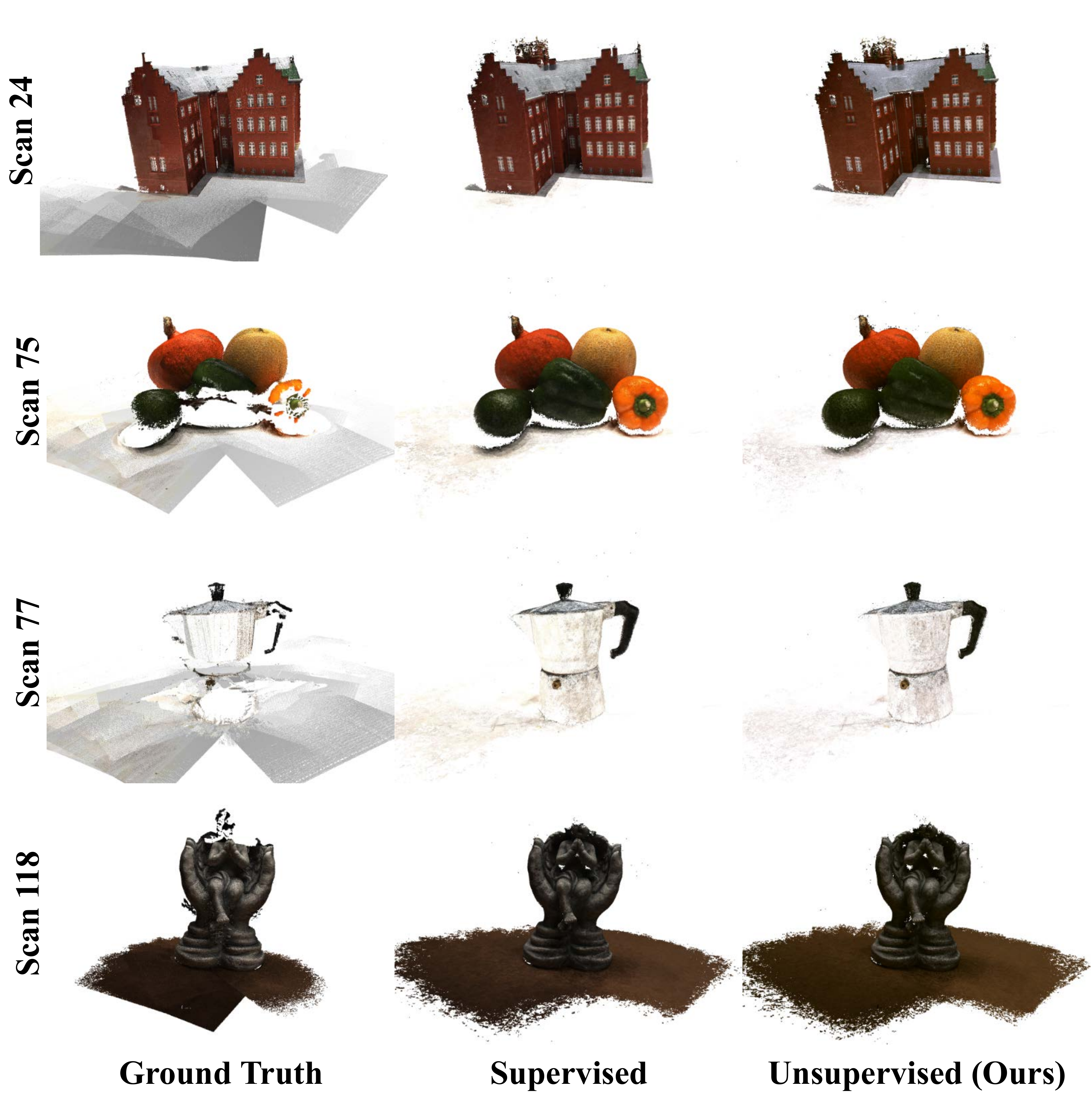}
\end{center}
    \vspace{-0.3cm}
    \caption{Qualitative comparison of the 3D reconstruction results on DTU evaluation benchmark. From left to right: Ground truth, results of SOTA supervised method, and our unsupervised method. CascadeMVSNet \cite{gu2020cascade} is utilized as the backbone.}
    \vspace{-0.3cm}
\label{fig9}
\end{figure}

\noindent\textbf{Error Metrics:}
In the DTU benchmark, \emph{Accuracy} is measured as the distance from the result to the structured light reference, encapsulating the quality of reconstruction; \emph{Completeness} is measured as the distance from the ground truth reference to the reconstructed result, encapsulating how much of the surface is captured; 
\emph{Overall} is a the average of \emph{Accuracy} and \emph{Completeness}, acting as a composite error metric.
In the Tanks\&Temples benchmark, F-score in each scene is calculated following the official evaluation process.


\noindent\textbf{Implementation Details:}
The backbone of our U-MVS framework is inherited from the consice open implementations of MVSNet \cite{yao2018mvsnet} and CascadeMVSNet \cite{gu2020cascade}.
In the preparation phase, we utilize a self-supervised method \cite{liu2020learning} to train an optical flow estimation network, PWC-Net \cite{sun2018pwc}, from the scratch on DTU dataset.
The two-view pairs for optical flow estimation are selected by combining the reference view with each of the source views provided by MVSNet \cite{yao2018mvsnet}.
Then, we utilize the self-supervised pretrained PWC-Net to estimate the optical flow from the aforementioned two-view pairs in the RGB2Flow module.
More implementation details are provided in the supplementary materials.

\subsection{Benchmark Results on DTU}
\label{sec:experiment:implementation_details:dtu}

\noindent\textbf{Comparison with SOTA}:
To evaluate the performance of our proposed method, the quantitative results on the evaluation set of DTU benchmark \cite{aanaes2016large} are presented in Table \ref{tab1}.
In the table, state-of-the-art (SOTA) supervised and unsupervised methods are compared.
From the figure, we can find that our proposed method performs better than previous unsupervised method.
Under the error metric of \emph{overall} in DTU benchmark, the performance of current SOTA supervised methods is about 0.351 - 0.355.
Whereas, without utilizing any ground truth labels, our unsupervised model with a backbone of CascadeMVSNet can achieve 0.3537 on \emph{overall} metric, which is comparable with supervised components.
Fig. \ref{fig9} shows the qualitative comparisons of the 3D reconstruction results on several scenes of DTU evaluation set.
With the same CascadeMVSNet as backbone, our self-supervision framework can achieve a comparable performance with the supervised training.

\noindent\textbf{Supervised vs Self-supervised}:
To provide a fair comparison with the same backbone, we compare our proposed self-supervised MVS framework with the supervised training methods on MVSNet and CascadeMVSNet.
The performance of supervised baselines are taken from previous papers (MVSNet \cite{yao2018mvsnet}, CascadeMVSNet \cite{gu2020cascade}).
From the italics in Table \ref{tab1}, it demonstrates that our self-supervised framework can perform slightly better than its supervised counterpart in an equal setting.


\begin{table}
    \small
    \centering
    \begin{tabular}{ccc|ccc}
    \hline
        $L_{pc}$ & $L_{fc}$ & $L_{uc}$ & Acc. & Comp. & Overall \\ \hline \hline
        $\checkmark$ &  &  & 0.5527 & 0.5345 & 0.5436 \\ \hline
        $\checkmark$ & $\checkmark$ &   & 0.5063 & 0.4576 & 0.4820 \\ \hline
        $\checkmark$ & $\checkmark$ & $\checkmark$ & \textbf{0.4695} & \textbf{0.4308} & \textbf{0.4501} \\ \hline
    \end{tabular}
\caption{Ablation study of different components of our proposed self-supervision framework using MVSNet as backbone.}
\vspace{-0.3cm}
\label{tab3}
\end{table}

\begin{table}
    \small
    \centering
    \begin{tabular}{ccc|ccc}
    \hline
        $L_{pc}$ & $L_{fc}$ & $L_{uc}$ & Acc. & Comp. & Overall \\ \hline \hline
        $\checkmark$ &  &  & 0.4442 & 0.3641 & 0.4041 \\ \hline
        $\checkmark$ & $\checkmark$ &  & 0.3745 & 0.3833 & 0.3789 \\ \hline
        $\checkmark$ & $\checkmark$ & $\checkmark$ & \textbf{0.3540} & \textbf{0.3535} & \textbf{0.3537} \\ \hline
    \end{tabular}
\caption{Ablation study of different components of our proposed self-supervision framework using CasMVSNet as backbone.}
\vspace{-0.3cm}
\label{tab4}
\end{table}

\noindent\textbf{Ablation Studies}:
To evaluate the effect of different self-supervised components in the proposed framework, we respectively train the model with different combinations of the self-supervised losses.
With a MVSNet as backbone, the quantitative results are presented in Table \ref{tab3}.
With a CascadeMVSNet as backbone ,the ablation results are presented in Table \ref{tab4}.
$L_{pc}$, $L_{fc}$, $L_{uc}$ represent the basic photometric consistency loss (Eq. \ref{eq3}), flow-depth consistency loss (Eq. \ref{eq8}), uncertainty-aware self-training consistency loss (Eq. \ref{eq12}) respectively.
From the tables, we can find that these self-supervised components can effectively improve the performance on all metrics.

\begin{figure}[t]
\begin{center}
    \includegraphics[width=\linewidth]{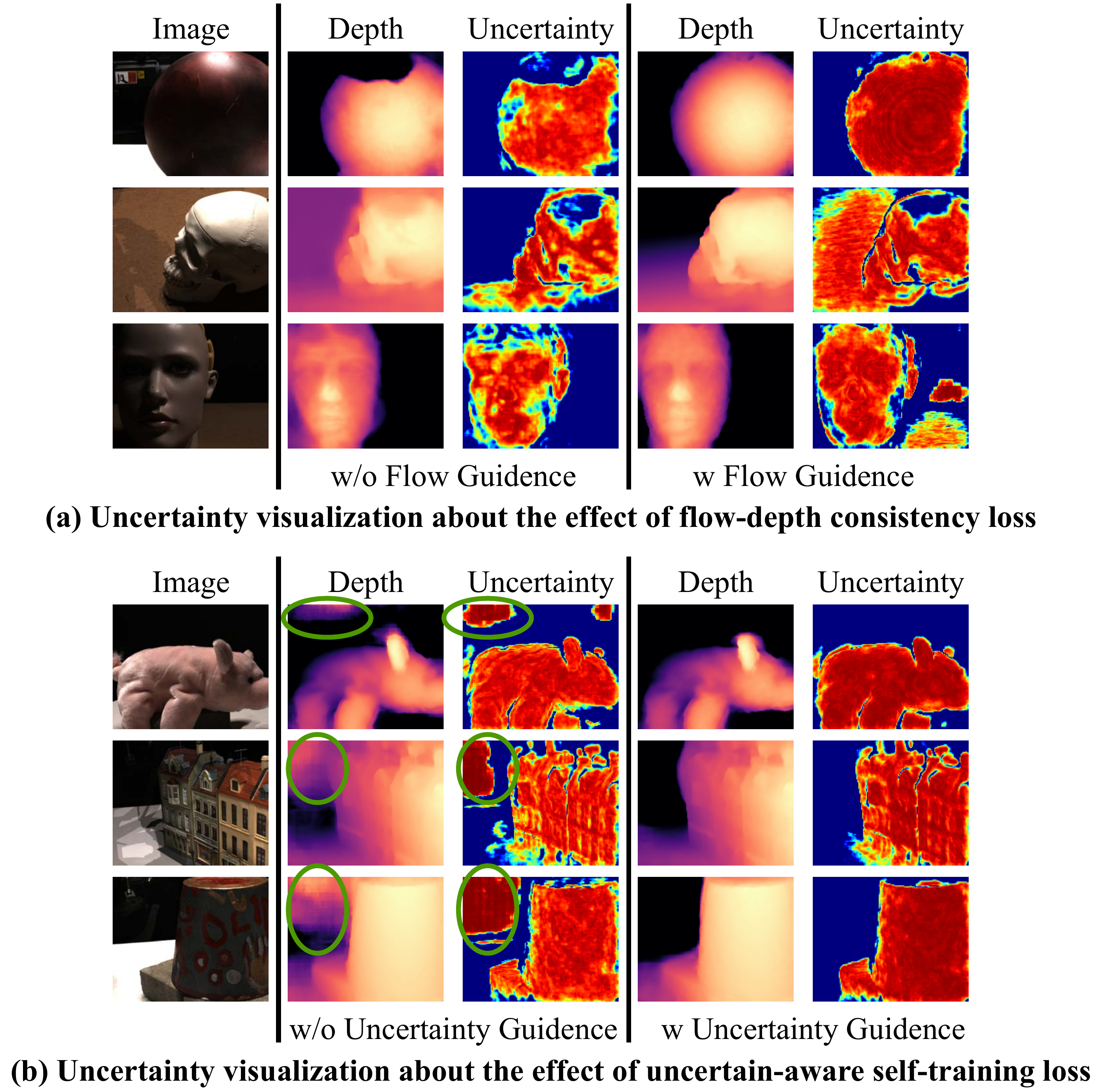}
\end{center}
\vspace{-0.2cm}
\caption{Visualization results of the uncertainty under the effect of our proposed flow-depth consistency loss $L_{fc}$ (Eq. \ref{eq4}) and uncertainty-aware self-training loss $L_{uc}$ (Eq. \ref{eq12}).}
\vspace{-0.2cm}
\label{fig-unc-viz}
\end{figure}

\begin{table*}[t]
    \small
    \centering
    \begin{tabular}{l|c|c|p{20pt}p{20pt}p{20pt}c p{20pt}c c p{20pt}}
    \hline
        Method & Sup. & Mean & Family & Francis & Horse & Lighthouse & M60 & Panther & Playground & Train \\ \hline \hline
        OpenMVG \cite{moulonopenmvg} + MVE \cite{fuhrmann2014mve} & - & 38.00 & 49.91 & 28.19 & 20.75 & 43.35 & 44.51 & 44.76 & 36.58 & 35.95 \\
        OpenMVG \cite{moulonopenmvg} + OpenMVS \cite{openmvs2020} & - & 41.71 & 58.86 & 32.59 & 26.25 & 43.12 & 44.73 & 46.85 & 45.97 & 35.27 \\
        COLMAP \cite{schonberger2016structure} & - & 42.14 & 50.41 & 22.25 & 25.63 & \textbf{56.43} & 44.83 & 46.97 & 48.53 & 42.04 \\
        \hline
        MVSNet \cite{yao2018mvsnet} & $\checkmark$ & 43.48 & 55.99 & 28.55 & 25.07 & 50.79 & 53.96 & 50.86 & 47.90 & 34.69 \\ 
        CIDER \cite{xu2020learning} & $\checkmark$ & 46.76 & 56.79 & 32.39 & 29.89 & 54.67 & 53.46 & 53.51 & 50.48 & 42.85 \\
        R-MVSNet \cite{yao2019recurrent} & $\checkmark$ & 48.40 & 69.96 & 46.65 & 32.59 & 42.95 & 51.88 	 & 48.80 & 52.00 & 42.38 \\
        CVP-MVSNet \cite{yang2020cost} & $\checkmark$ & 54.03 & 76.50 & 47.74 & 36.34 & 55.12 & \textbf{57.28} & \textbf{54.28} & 57.43 & 47.54 \\
        CascadeMVSNet \cite{gu2020cascade} & $\checkmark$ & 56.42 & 76.36 & 58.45 & 46.20 & 55.53 & 56.11 & 54.02 & \textbf{58.17} & 46.56 \\
        \hline
        MVS$^2$ \cite{dai2019mvs2} & $\times$ & 37.21 & 47.74 & 21.55 & 19.50 & 44.54 & 44.86 & 46.32 & 43.38 & 29.72 \\ 
        M$^3$VSNet \cite{huang2020m} & $\times$ & 37.67 & 47.74 & 24.38 & 18.74 & 44.42 & 43.45 & 44.95 & 47.39 & 30.31 \\ 
        JDACS \cite{xu2021self} & $\times$ & 45.48 & 66.62 & 38.25 & 36.11 & 46.12 & 46.66 & 45.25 & 47.69 & 37.16 \\
        \textbf{Ours + CascadeMVSNet} & $\times$ & \textbf{57.15} & \textbf{76.49} & \textbf{60.04} & \textbf{49.20} & 55.52 & 55.33 & 51.22 & 56.77 & \textbf{52.63} \\
        \hline
    \end{tabular}
\caption{Quantitative results on the \emph{intermediate} partition of \emph{Tanks and Temples} benchmark without any finetuning. We present the \emph{f-score} result of all submissions from the official leaderboard of \emph{Tanks and Temples} benchmark.}
\vspace{-0.2cm}
\label{tab5}
\end{table*}

\begin{table*}[t]
    \small
    \centering
    \begin{tabular}{l|c|c|cccccc}
    \hline
        Method & Sup. & Mean & Auditorium & Ballroom & Courtroom & Museum & Palace & Temple  \\ \hline \hline
        COLMAP \cite{schonberger2016structure} & - & 27.24 & 16.02 & 25.23 & 34.70 & 41.51 & 18.05 & 27.94  \\
        \hline
        R-MVSNet \cite{yao2019recurrent} & $\checkmark$ & 24.91 & 12.55 & 29.09 & 25.06 & 38.68 & 19.14 & 24.96 \\ 
        CIDER \cite{xu2020learning}  & $\checkmark$ & 23.12 & 12.77 & 24.94 & 25.01 & 33.64 & 19.18 & 23.15  \\ 
        CascadeMVSNet \cite{gu2020cascade} & $\checkmark$ & \textbf{31.12} & 19.81 & \textbf{38.46} & \textbf{29.10} & \textbf{43.87} & 27.36 & 28.11 \\
        \hline
        \textbf{Ours + CascadeMVSNet} & $\times$ & 30.97 & \textbf{22.79} & 35.39  & 28.90 & 36.70 & \textbf{28.77} & \textbf{33.25} \\
        \hline
    \end{tabular}
\caption{Quantitative results on the \emph{advanced} partition of \emph{Tanks and Temples} benchmark without any finetuning. We present the \emph{f-score} result of all submissions from the official leaderboard of \emph{Tanks and Temples} benchmark.}
\vspace{-0.2cm}
\label{tab6}
\end{table*}

\noindent\textbf{Uncertainty Visualization}:
To find out whether the proposed self-supervised components can handle the aforementioned issues of uncertainties in foreground and background in Sec. \ref{sec:introduction}, we provide the visualization results of the uncertainty estimated by Monte-Carlo Dropout in Fig. \ref{fig-unc-viz}.
For the first question, the uncertainty maps of the models respectively trained with or without our proposed flow-depth consistency loss $L_{fc}$ are presented in Fig. \ref{fig-unc-viz}(a).
With the guidance of the dense 2D correspondence in flow-depth consistency loss, it is found that the certain regions in self-supervision become larger and more certain.
It demonstrates that effective supervision towards disturbance such as reflection and low-texture is involved via the extra correspondence prior of flow-depth consistency.
For the second question, the uncertainty maps of the models respectively trained with or without uncertainty guidance in the self-training loss $L_{uc}$ are shown in Fig. \ref{fig-unc-viz}(b).
From the figure, we can find that if the model is trained without the guidance of uncertainty, the interfused uncertain supervision may be mistaken for correct pseudo label, further misleading the self-supervision.
With the guidance of uncertainty, the misleading effect is alleviated, as shown in Fig. \ref{fig-unc-viz}(b).
It shows that the proposed uncertainty-aware self-training loss can enhance the supervision signals and get rid of the negative effect of uncertain supervision signals in self-supervised MVS.

\begin{figure}[t]
\begin{center}
\includegraphics[width=0.8\linewidth]{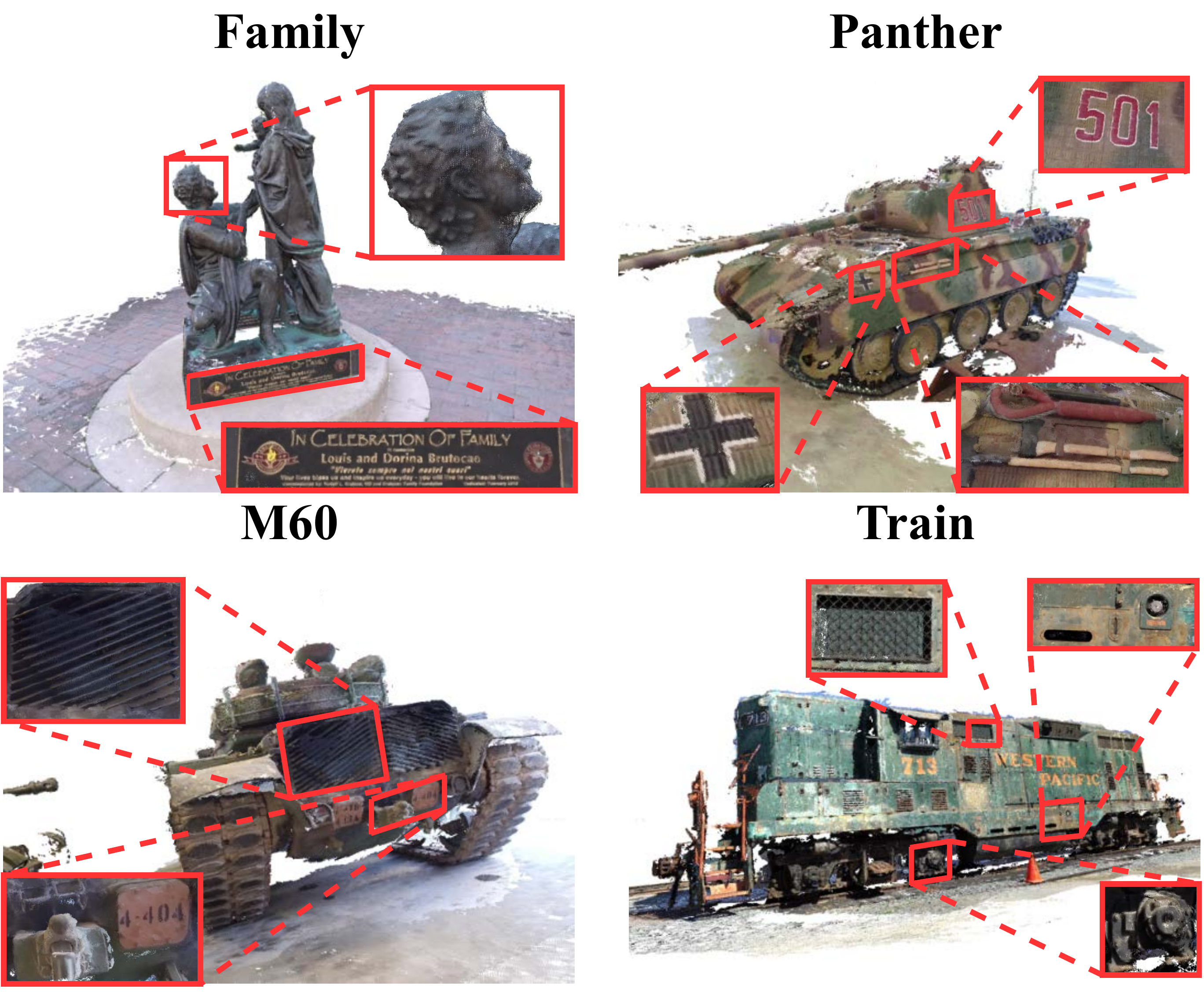}
\end{center}
   \vspace{-0.2cm}
   \caption{Visualization of the reconstructed 3D model on the \emph{intermediate} partition of \emph{Tanks and Temples} benchmark.}
   \vspace{-0.3cm}
\label{fig-tanks-visualization-intermediate}
\end{figure}

\subsection{Generalization}
\label{sec:experiment:generalization}

In order to evaluate the generalization ability of the proposed method, we compare the performance of our proposed method with state-of-the-art supervised and unsupervised methods on Tanks and Temples benchmark.
For a fair comparison, we utilize the model merely trained on DTU dataset without any finetuning to test on Tanks\&Temples dataset.
For evaluation, the input image is set to $1920 \times 1056$, and the number of views is 7.
We use CascadeMVSNet as backbone without using any ground truth in the training phase.
The quantitative comparisons of performance on the \emph{intermediate} partition of Tanks and Temples benchmark is presented in Table \ref{tab5}.
The experimental results in the table demonstrate that our proposed method has the highest score compared with unsupervised methods.
Furthermore, the mean F-score on the \emph{intermediate} benchmark is 57.15 which also outperforms previous supervised opponents including CascadeMVSNet.
On the more complex \emph{advanced} partition of Tanks and Temples benchmark, the comparison results are provided in Table \ref{tab6}.
Without using any ground truth annotations, our proposed method can still present comparable performance with the SOTA supervised methods.
The visualization results of the reconstructed 3D model on the \emph{intermediate} partition of \emph{Tanks and Temples} benchmark is provided in Fig. \ref{fig-tanks-visualization-intermediate}.
Our proposed method achieves the best performance among unsupervised MVS methods on both partitions of \emph{Tanks and Temples} benchmark until March 17, 2021.

\section{Conclusions}

In this paper, we have proposed a novel Uncertainty reduction Multi-view Stereo framework (U-MVS) for self-supervised learning, aiming to handle the two discovered problems via uncertainty analysis: 1) \emph{Ambiguous supervision in foreground}; 2) \emph{Invalid supervision in background}.
For the first problem, we propose a flow-depth consistency loss to endow dense 2D correspondence of optical flows to regularize the 3D stereo correspondence in self-supervised MVS.
For the second problem, we use Monte-Carlo Dropout to estimate the uncertainty map and filter the uncertain parts from supervision.
The experimental results demonstrate the effectiveness of our proposed U-MVS framework.

\section{Acknowledgement}

This work is partially supported by  National Natural Science Foundation of China (61876176, 61976095), the Science and Technology Service Network Initiative of Chinese Academy of Sciences(KFJ-STS-QYZX-092), Guangdong NSF Project (No. 2020B1515120085), the Shanghai Committee of Science and Technology, China (Grant No. 20DZ1100800 and 21DZ1100100). This work is supported by Alibaba Group through Alibaba Innovative Research (AIR) Program.

\clearpage

\section{Appendix}

\subsection{Why Use Uncertainty for Self-supervised MVS?}

\begin{figure}[t]
\begin{center}
   \includegraphics[width=\linewidth]{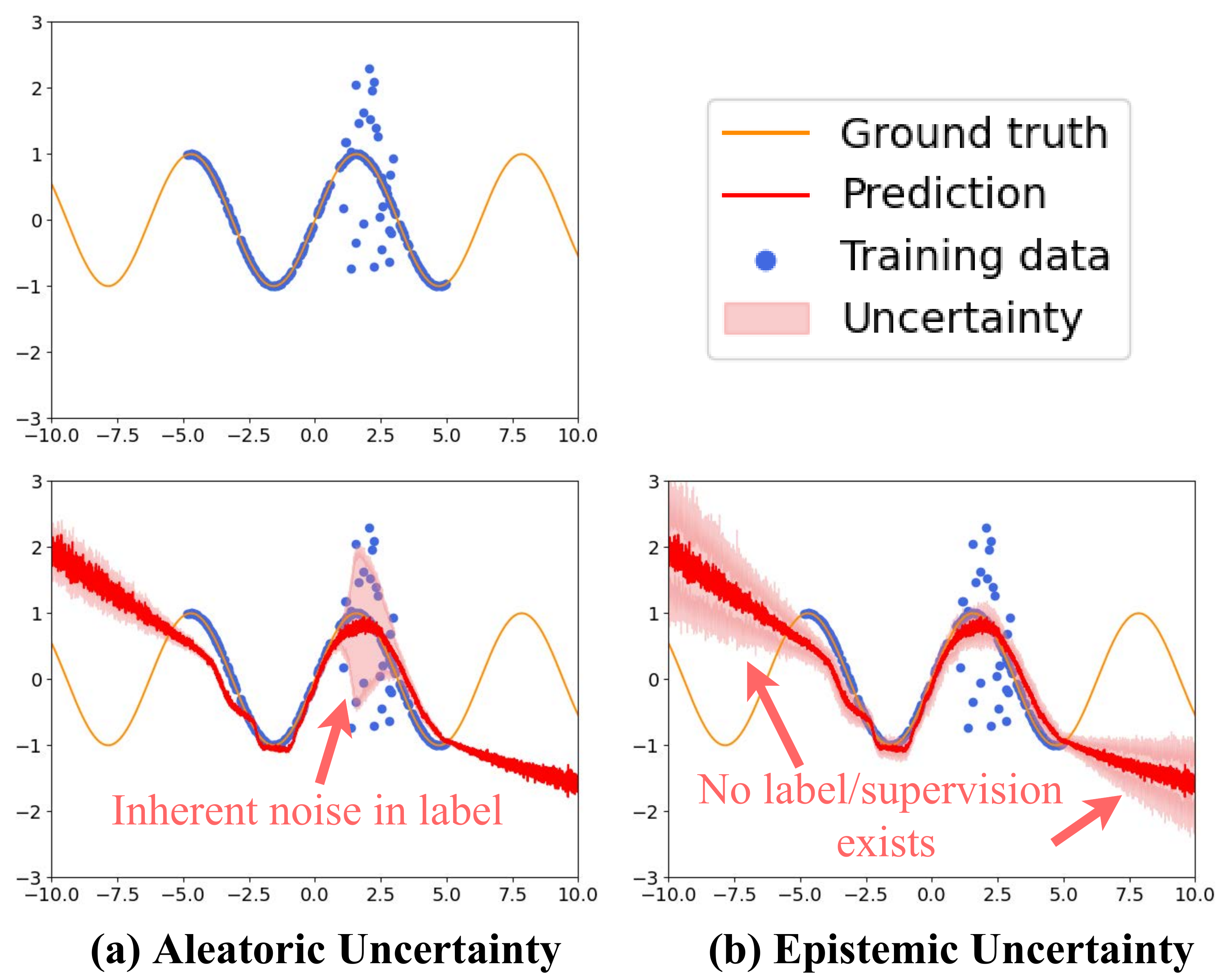}
\end{center}
\vspace{-0.2cm}
   \caption{A toy example to understand aleatoric and epistemic uncertainty.}
\vspace{-0.2cm}
\label{fig:toy_example}
\end{figure}

In Bayesian deep learning, the uncertainty is categorized into two types \cite{der2009aleatory}: aleatoric and epistemic uncertainty.
Aleatoric uncertainty models the inherent noise in the training data, and epistemic uncertainty accounts for what is not included in the training data.
As shown in Fig. \ref{fig:toy_example}, a toy example of aleatoric and epistemic uncertainty is provided.
In Fig. \ref{fig:toy_example}(a), aleatoric uncertainty models the regions which have noisy labels.
In Fig. \ref{fig:toy_example}(b), it shows that epistemic uncertainty models what current model ignores, for example, the regions without certain supervision or label.

In previous works \cite{khot2019learning, dai2019mvs2, huang2020m, xu2021self}, self-supervised MVS methods are built on intuitive assumptions, aiming at involving extra priors into the self-supervision loss.
It can be further viewed as an attempt to increase the certain supervision signals in self-supervision, which is proved by extensive experiments to be effective.
Whereas, from an opposite viewpoint, we rethink the effect of uncertain supervision signals modeled by \emph{epistemic uncertainty} in this work.
Since epistemic uncertainty models the ignorance of supervision, it can provide us a more comprehensive and explainable understanding of self-supervision.
In analogy to the epistemic uncertainty in Fig. \ref{fig:toy_example}(b), the uncertainty in self-supervision can guide our skepticism to the limitations of current self-supervised MVS, which is further discussed in the Introduction part of the manuscript.

\subsection{Modified Backbone Network for Uncertainty Estimation}

\begin{figure*}[t]
\begin{center}
   \includegraphics[width=0.8\linewidth]{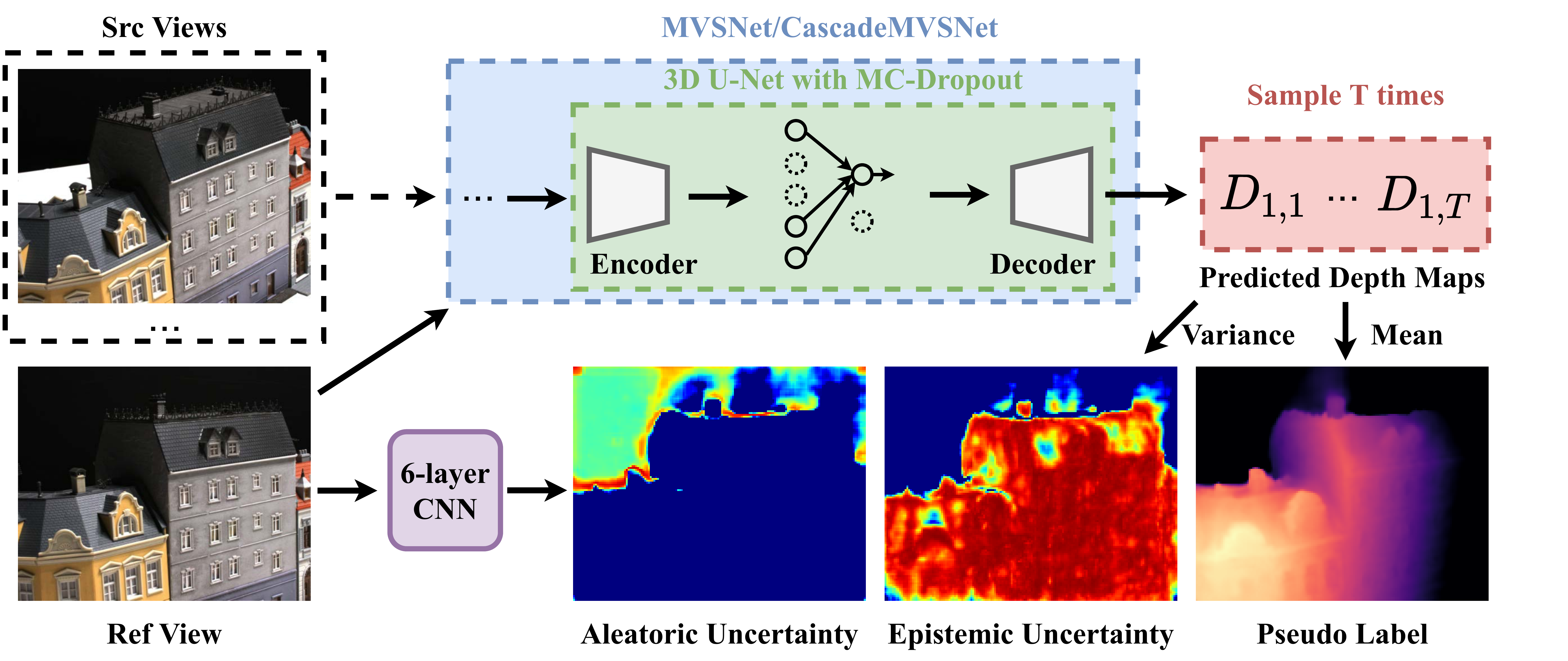}
\end{center}
\vspace{-0.2cm}
   \caption{Illustration of the modified architecture of backbone network for estimation aleatoric uncertainty and epistemic uncertainty.}
\vspace{-0.2cm}
\label{fig:backbone_architecture}
\end{figure*}

In this section, we introduce the modified backbone network for estimating the aforementioned aleatoric uncertainty and epistemic uncertainty, following a classical configuration proposed by \cite{kendall2017uncertainties}.
As shown in Fig. \ref{fig:backbone_architecture}, the illustration of the modified backbone architecture is presented.
The aleatoric uncertainty map is directly predicted by a 6-layer CNN, whose detailed architecture is further listed in Table \ref{tab:network_aleatoric}.
The epistemic uncertainty map is estimated via a statistical Bayesian model by sampling $T$ times.
Though traditional Bayesian models can offer a mathematically grounded framework to estimate model uncertainty, they are usually attached with prohibitive computational cost.
Hence, Monte-Carlo Dropout (MC-Dropout) \cite{gal2016dropout} attempts to alleviate the huge cost in computation, casting dropout training in deep neural networks as approximate Bayesian inference in deep Gaussian process.
In Monte-Carlo Dropout, the inference is done by training a model with dropout, and by also performing dropout at test time to sample from the approximate posterior.
Detailed theoretical evidence is also provided in \cite{gal2016dropout}.
Following an open implementation\footnote{\url{https://github.com/pmorerio/dl-uncertainty}} of \cite{kendall2017uncertainties}, we append dropout layers on the bottleneck layers of the 3D U-Net in MVSNet \cite{yao2018mvsnet} and CascadeMVSNet \cite{gu2020cascade}, which are applied as backbone in our proposed self-supervised MVS framework.
If MVSNet is applied as backbone, the MC-Dropout is embedded in the 3D U-Net of MVSNet, whose details are provided in Table \ref{tab:network_structure_mc_dropout}.
If CascadeMVSNet is applied as backbone, the modified 3D U-Net with MC-Dropout shares the same architecture as Table \ref{tab:network_structure_mc_dropout} shows.
Since CascadeMVSNet has multiple stages, the MC-Dropout layers are only activated on the first stage.
Because too many dropout layers may result into strong regularization effect in self-supervised training, and make the model trapped in a trivial solution.
To guarantee the convergence of self-supervised training in MVS, the number of dropout layers is limited.

\begin{table}[t]
\small
\centering
\begin{tabular}{l|l|l}
\hline
Name    & Layer                  & Output Size \\ \hline \hline
Input   & -                      & H$\times$W$\times$3       \\
Conv\_0 & ConvBR,K=3$\times$3,S=1,F=16  & H$\times$W$\times$16      \\
Conv\_1 & ConvBR,K=3$\times$3,S=1,F=32  & H$\times$W$\times$32      \\
Conv\_2 & ConvBR,K=3$\times$3,S=1,F=128 & H$\times$W$\times$128     \\
Conv\_3 & ConvBR,K=3$\times$3,S=1,F=256 & H$\times$W$\times$256     \\
Conv\_4 & ConvBR,K=3$\times$3,S=1,F=32  & H$\times$W$\times$32      \\
Conv\_5 & Conv,K=3$\times$3,S=1,F=1     & H$\times$W$\times$1       \\
\hline
\end{tabular}
\caption{Network structure of the 6-layer CNN utilized to estimate \emph{aleatoric uncertainty}. Denote that 3D convolution as ``Conv'', 3D deconvolution as ``DeConv'', batch normalization as ``B'', ReLU as ``R'' in the column of ``Layer''. ``K'' is the kernel size, ``S'' is the stride and ``F'' is the output channels. ``H'' and ``W'' represent the height and width, respectively.}
\label{tab:network_aleatoric}
\end{table}

\begin{table}[t]
\small
\centering
\begin{tabular}{l|l|l}
\hline
Name         & Layer                   & Output Size \\ \hline \hline
Input        & -                       & H$\times$W$\times$32      \\
Conv\_0      & ConvBR,K=3$\times$3,S=1,F=8    & H$\times$W$\times$8       \\
Conv\_1      & ConvBR,K=3$\times$3,S=2,F=16   & H/2$\times$W/2$\times$16  \\
Conv\_2      & ConvBR,K=3$\times$3,S=1,F=16   & H/2$\times$W/2$\times$16  \\
Conv\_3      & ConvBR,K=3$\times$3,S=2,F=32   & H/4$\times$W/4$\times$32  \\
Conv\_4      & ConvBR,K=3$\times$3,S=1,F=32   & H/4$\times$W/4$\times$32  \\
Conv\_5      & ConvBR,K=3$\times$3,S=2,F=64   & H/8$\times$W/8$\times$64  \\
Conv\_6      & ConvBR,K=3$\times$3,S=1,F=64   & H/8$\times$W/8$\times$64  \\
\color{red}{Drop\_6}      & Dropout,Rate=0.5         & H/8$\times$W/8$\times$64  \\
Deconv\_7    & DeConvBR,K=3$\times$3,S=2,F=32 & H/4$\times$W/4$\times$32  \\
\color{red}{Drop\_7}      & Dropout,Rate=0.5         & H/4$\times$W/4$\times$32  \\
Shortcut\_8  & Deconv\_7 + Conv4       & H/4$\times$W/4$\times$32  \\
Deconv\_9    & DeConvBR,K=3$\times$3,S=2,F=16 & H/2$\times$W/2$\times$16  \\
Shortcut\_10 & Deconv\_9 + Conv2       & H/2$\times$W/2$\times$16  \\
Deconv\_11   & DeConvBR,K=3$\times$3,S=2,F=8  & H$\times$W$\times$8       \\
Shortcut\_12 & Deconv\_11 + conv0      & H$\times$W$\times$8       \\
Conv\_13     & Conv,K=3$\times$3,S=1,F=1      & H$\times$W$\times$1       \\
\hline
\end{tabular}
\caption{Network structure of modified 3D U-Net in MVSNet, embeded with \emph{Monte-Carlo Dropout} to estimate \emph{epistemic uncertainty}. Denote that 3D convolution as ``Conv'', 3D deconvolution as ``DeConv'', batch normalization as ``B'', ReLU as ``R'' in the column of ``Layer''. ``K'' is the kernel size, ``S'' is the stride, ``F'' is the output channels and ``Rate'' means the dropout rate. ``H'' and ``W'' represent the height and width, respectively.}
\label{tab:network_structure_mc_dropout}
\end{table}

\subsection{Implementation Details}

\textbf{Backbone}: We directly adopt the concise open implementations of MVSNet\footnote{\url{https://github.com/xy-guo/MVSNet_pytorch}} and CascadeMVSNet\footnote{\url{https://github.com/alibaba/cascade-stereo/tree/master/CasMVSNet}} as the bacbone in our porposed U-MVS framework.
Following the suggestions proposed by \cite{kendall2017uncertainties}, the backbone architecture is embedded with MC-Dropout and a 6-layer CNN for uncertainty estimation, which is introduced in the previous section.
In default, the other network settings follow the original open implementation.

\begin{table*}[t]
\centering
\begin{tabular}{c|ccc|ccc|ccc}
\hline
    & \multicolumn{3}{c|}{DTU}          & \multicolumn{3}{c}{Intermediate of Tanks\&Temples} & \multicolumn{3}{|c}{Advanced of Tanks\&Temples} \\ \hline \hline
Sup & Accuracy & Completeness & Overall & Precision     & Recall     & F-score     & Precision    & Recall    & F-score    \\ \hline 
   $\checkmark$ & \color{red}{0.325}    & 0.385       & 0.355  & \color{red}{47.62}         & 74.01      & 56.84       & \color{red}{29.68}        & 35.24     & \color{red}{31.12}      \\
   $\times$ & 0.354    & \color{red}{0.3535}       & \color{red}{0.3537}  & 45.45         & \color{red}{78.52}      & \color{red}{57.15}       & 24.22        & \color{red}{44.46}     & 30.97      \\ \hline
\end{tabular}
\caption{Performance comparison of our self-supervised method and supervised method on DTU evaluation set, intermediate and advanced partition set of Tanks\&Temples. CascadeMVSNet \cite{gu2020cascade} is utilized as the backbone. Under the metrics of DTU benchmark, the smaller the value the better the performance; Under the metrics of Tanks\&Temples, the larger the value the better the performance.}
\label{tab:analysis_sup_self-sup}
\end{table*}

\textbf{RGB2Flow Module}: In the RGB2Flow module, we utilize a self-supervised method\footnote{\url{https://github.com/lliuz/ARFlow}} to train an optical flow estimation network, PWC-Net \cite{sun2018pwc}, from the scratch on DTU dataset.
The two-view pairs for estimating optical flow are selected by combining the reference view with each of the source views in the multi-view pairs provided by MVSNet \cite{yao2018mvsnet}.
After self-supervisedly pretraining the PWC-Net, it is able to predict the optical flow from RGB images in the RGB2Flow module.
No extra ground truth is used in this module.

\textbf{Uncertainty Estimation}: As discussed in the manuscript, MC-Dropout is only activated during estimating the uncertainty maps.
We proceed the forward propagation on the network with random MC-Dropout for $T=20$ times, which can be viewed as sampling $T$ different model weights.
Following the procedure of \cite{kendall2017uncertainties}, the mean and variance of $T$ sampled depth maps are respectively treated as the pseudo label and uncertainty map, as shown in Fig. \ref{fig:backbone_architecture}.

\textbf{Training and testing}: The whole training process is conducted on 4 RTX 2080Ti GPU.
No ground truth depth maps are used in the training phase\footnote{The code will be released on Github in the future}.
In default, the hyperparameter settings for self-supervision follow the  bconfiguration of Unsup\_MVS \cite{khot2019learning}.
In the training phase, the image resolution is set to $640 \times 512$.
Due to the limitation of memory, the batch size is set to 1 per GPU.
The model is trained on the DTU training set as \cite{yao2018mvsnet}.
If MVSNet is selected as backbone, the model is firstly trained for 10 epochs in the self-supervision pretraining stage, and the model is further trained for 10 epochs in the pseudo label post-training stage.
If CascadeMVSNet is selected as backbone, the self-supervision pretraining stage requires 16 epochs for training the model, and the pseudo label post-training stage requires 16 epochs for further training.
We utilize Adam optimizer with a learning rate of 1e-3 which is decreased by 0.5 times every 2 epochs.
In the testing phase, the depth maps on all views of the scene are predicted.
After depth estimation, 3D point cloud is reconstructed from the multi-view depth maps and images \cite{galliani2015massively}.
The test setting is also the same as the aforementioned open implementations.

\subsection{Discussion}

\subsubsection{Quantitative evaluation of uncertainty estimates.}

\begin{figure}[t]
\begin{center}
   \includegraphics[width=\linewidth]{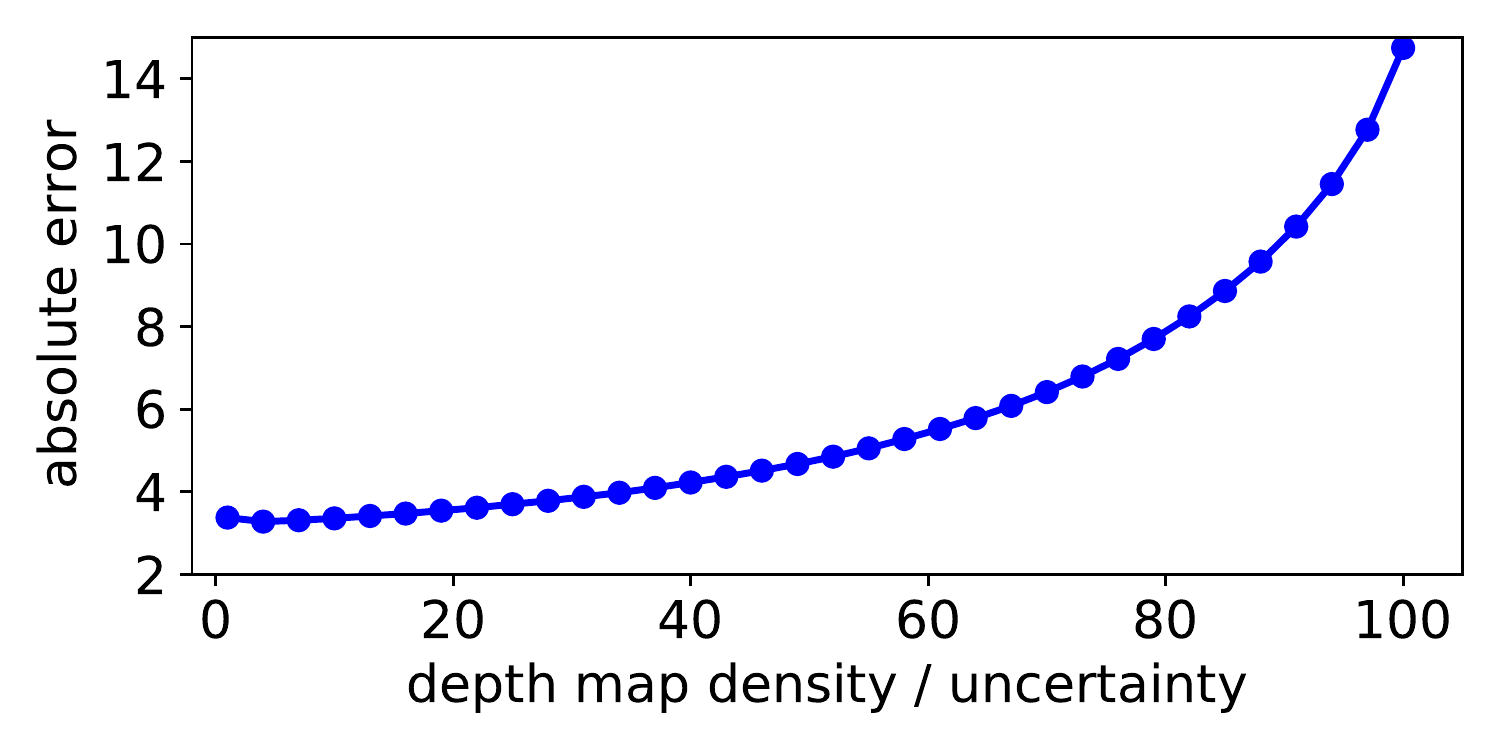}
\end{center}
\vspace{-0.2cm}
   \caption{Quantitative evaluation results of uncertainty estimates.}
\vspace{-0.2cm}
\label{fig-uncertainty-quantitative-results}
\end{figure}

Ideally, the aforementioned uncertainty should be inversely correlated with accuracy.
To provide a quantitative evaluation of uncertainty estimates, we provide further experimental results in Fig. \ref{fig-uncertainty-quantitative-results} following \cite{hu2012quantitative}.
As suggested by the authors, in order to assess the capability of the uncertainty measure to predict whether a prediction is (in)correct, the depth predictions on all pixels are ranked in decreasing order of confidence.
Then the per-pixel error rate of the depth predictions are computed.
As shown in the figure, the abscissa represented the percentage of selected pixels ranked by the uncertainty, which is also called ``density''\cite{hu2012quantitative}.
The ordinate in the figure shows the absolute error rate of the selected pixels according its density.
It is noted that as the density/uncertainty increases, the absolute error rate increases as well.
We can find that the aforementioned uncertainty is inversely correlated with accuracy, which supports the idea of rejecting invalid supervisions according to uncertainty estimates. 

\subsubsection{Can Self-supervised Methods ``Outperform'' Supervised Methods?}
\label{sec:appendix:self_sup_sup}

In Table. \ref{tab:analysis_sup_self-sup}, we provide a direct comparison of our proposed self-supervised MVS framework and supervised method on several benchmarks with the same backbone of CascadeMVSNet \cite{gu2020cascade}, such as DTU, intermediate and advanced partition in Tanks\&Temples.
On the third row of Table \ref{tab:analysis_sup_self-sup}, the performance of supervised method is picked from the original paper and the official leaderboard of Tanks\&Temples.
On the fourth row of Table \ref{tab:analysis_sup_self-sup}, we provide the performance of our proposed self-supervised method under the same metrics.
The performance of our proposed unsupervised method on intermediate\footnote{The submission is named as \emph{6956-ss-mvs-test} on \url{https://www.tanksandtemples.org/leaderboard/IntermediateF/}} and advanced\footnote{The submission is named as \emph{6956-self-sup-mvs} on \url{https://www.tanksandtemples.org/leaderboard/AdvancedF/}} partition of Tanks\&Temples benchmark can be found on the official website.

From the quantitative comparison in Table \ref{tab:analysis_sup_self-sup}, we can find: the Accuracy/Precison of supervised method is better than our unsupervised method; Whereas, the Completeness/Recall of self-supervised method is better than the supervised one; Moreover, the Overall/F-score of self-supervised method is competitive with the supervised method.
It demonstrates that each of the supervised and self-supervised methods has its own advantages.
For supervised training, the results are more precise and accurate on each point of the reconstructed 3D point cloud, compared with self-supervised method.
Because there is an inevitable loss of detailed information in self-supervision signal built upon cross-view correspondence between discrete pixels.
For our self-supervised MVS framework, the advantage is that the reconstructed 3D model can retain more integral parts, compared with supervised training.
It shows that our self-supervised framework can excavate depth information from abundant correspondence priors from the multi-view images which can cover more integral parts of the 3D object.

\subsubsection{Visualization of the Reconstructed 3D Models}

\begin{figure*}
\begin{center}
\includegraphics[width=\linewidth]{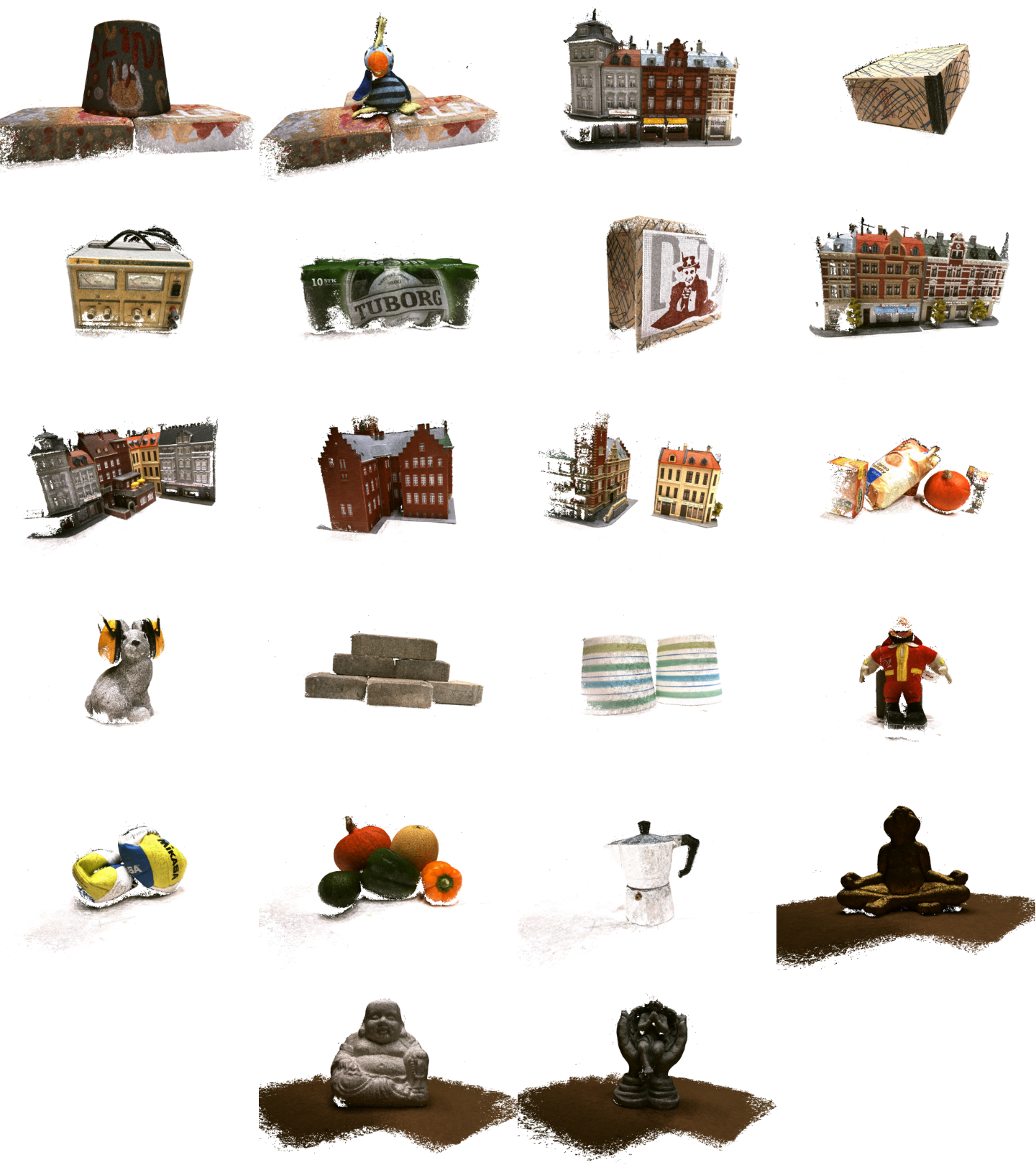}
\end{center}
   \caption{Visualization of all scenes in DTU evaluation set.}
\label{fig:dtu_all_viz}
\end{figure*}

\begin{figure*}
\begin{center}
\includegraphics[width=\linewidth]{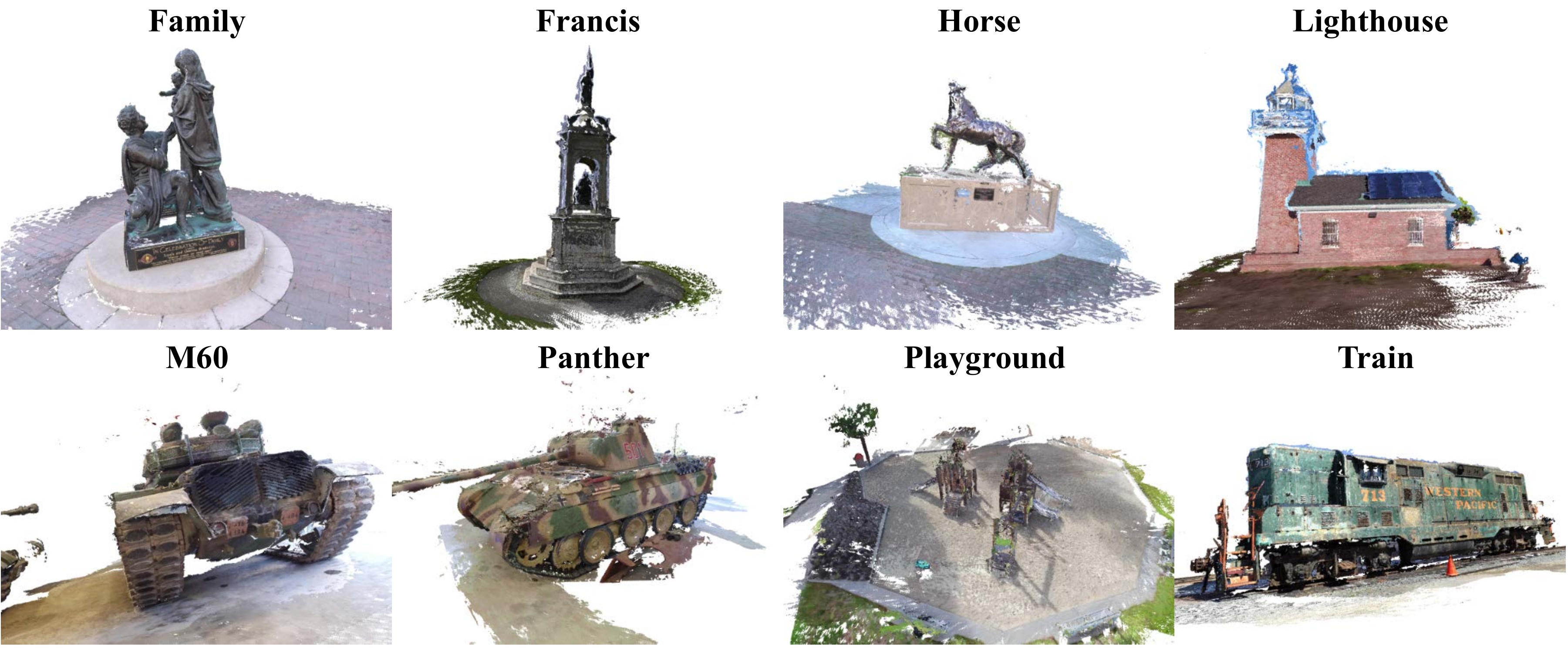}
\end{center}
   \caption{Visualization of all scenes on the \emph{intermediate} partition of Tanks\&Temples dataset.}
\label{fig:tanks_all_viz_intermediate}
\end{figure*}

\begin{figure*}
\begin{center}
\includegraphics[width=\linewidth]{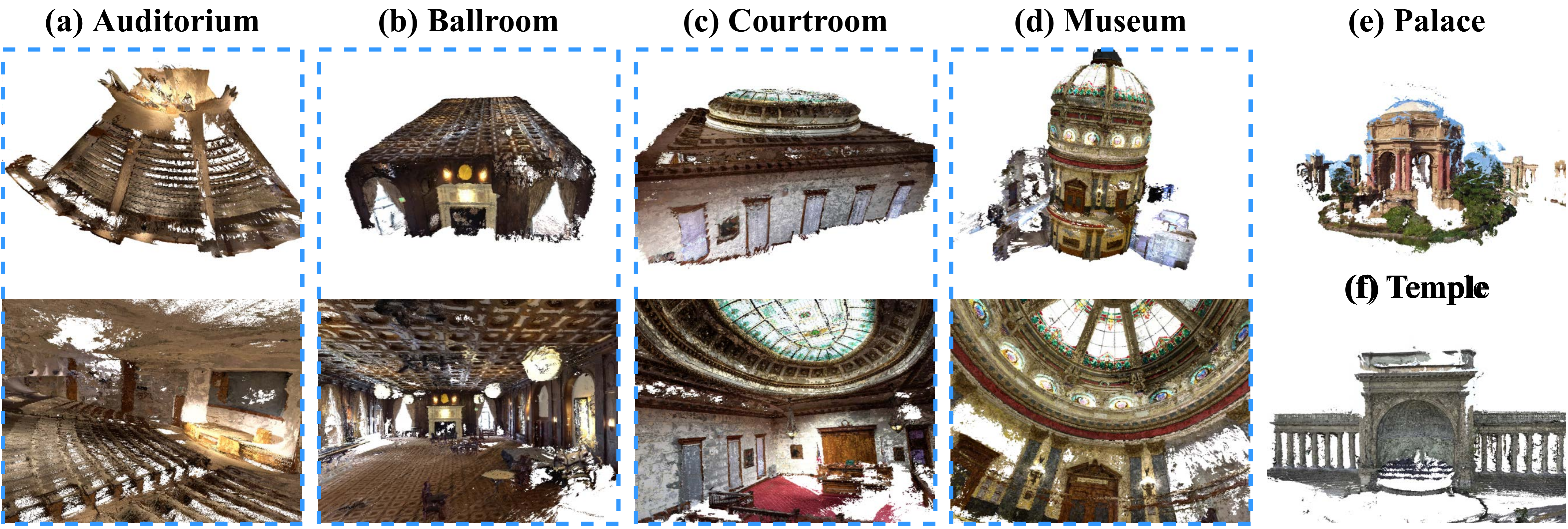}
\end{center}
   \caption{Visualization of all scenes on the \emph{advanced} partition of Tanks\&Temples dataset.}
\label{fig:tanks_all_viz_advanced}
\end{figure*}

We visualize the reconstructed 3D point clouds from DTU evaluation set and Tanks\&Temples test set respectively in Fig. \ref{fig:dtu_all_viz}, Fig. \ref{fig:tanks_all_viz_intermediate} and Fig. \ref{fig:tanks_all_viz_advanced}.

\section{Limitations}

\textbf{1) The computation and time consumption for uncertainty estimation is enormous.}
Though the MC-Dropout can alleviate the prohibitive computational cost of Bayesian model during estimating the uncertainty maps, it still requires to sample $T=20$ times, which means the model is forward propagated for 20 more times. 
It shows that the uncertainty estimation may be 20 times slower than a normal forward propagation.
As a result, the uncertainty estimation phase of our proposed U-MVS framework may be time-consuming.
In future work, a fast and light-weight method for uncertainty estimation is necessary.

\textbf{2) Extension in a semi-supervised framework is ignored.}
As discussed in Section \ref{sec:appendix:self_sup_sup} and Table \ref{tab:analysis_sup_self-sup}, each of the supervised method and our proposed self-supervised framework has its own advantage: the supervised training can reconstruct more accurate details of 3D model, whereas self-supervised method can retain integral parts in 3D reconstruction, producing more complete results.
The combination of these two methods may provide further improvements in the performance of 3D reconstruction, such as semi-supervised learning, in future works.

{
\bibliographystyle{ieee_fullname}
\bibliography{reference}
}

\end{document}